%% file: short_main.tex
\documentclass[11pt]{article}
\usepackage{acl}
\usepackage[most]{tcolorbox}
\usepackage{enumitem}
\usepackage{xcolor}
\usepackage{listings}
\usepackage{caption}
\usepackage{needspace}
\usepackage{ragged2e}
\usepackage{times}
\usepackage{latexsym}
\usepackage[T1]{fontenc}
\usepackage[utf8]{inputenc}
\usepackage{microtype}
\usepackage{inconsolata}
\usepackage{graphicx}
\usepackage{booktabs}
\usepackage{amsmath}
\usepackage{amssymb}
\usepackage{dblfloatfix}

\graphicspath{{figures/}}

\lstdefinestyle{jsonstyle}{ 
  basicstyle=\ttfamily\tiny, 
  breaklines=true, 
  breakatwhitespace=false, 
  columns=fullflexible, 
  keepspaces=true, 
  showstringspaces=false, 
  frame=single, 
  rulecolor=\color{black!25}, 
  xleftmargin=0pt, 
  xrightmargin=0pt 
}

\tcbset{ 
  verifierpromptbox/.style={ 
    enhanced, 
    breakable, 
    colback=white, 
    colframe=black!70, 
    boxrule=0.5pt, 
    arc=1pt, 
    left=6pt, 
    right=6pt, 
    top=6pt, 
    bottom=6pt, 
    width=\textwidth, 
    before upper={ \scriptsize \RaggedRight \setlength{\parindent}{0pt} \setlength{\parskip}{2pt} } 
  } 
}

\title{ChemCLIR-Bench: Benchmarking Cross-Lingual Information Retrieval in Multilingual Chemical Patents}

\author{
  Mahdi Astaraki\textsuperscript{\rm 1, 2}\thanks{\ \ Equal contribution.}\hspace{0.4em}%
  Mohammad Khodadad\textsuperscript{\rm 2}\footnotemark[1]\hspace{0.4em}%
  Reza Namazi\textsuperscript{\rm 2}\hspace{0.4em}%
  Mohammad Arshi Saloot\textsuperscript{\rm 2}\\
  \textbf{Amir Reza Behzad Moghadam}\textsuperscript{\rm 1} \quad
  \textbf{Hamidreza Mahyar}\textsuperscript{\rm 1} \quad
  \textbf{Soheila Samiee}\textsuperscript{\rm 2}\thanks{\ \ Corresponding author.}\\
  \textsuperscript{\rm 1}Faculty of Engineering, McMaster University, Canada\\
  \textsuperscript{\rm 2}BASF Canada Inc., Canada\\
  \texttt{\{astarakm, behzadma, mahyarh\}@mcmaster.ca}\\
  \texttt{\{mohammad.khodadad, reza.namazi\}@basf.com}\\
  \texttt{\{mohammad.arshi-saloot, soheila.samiee\}@basf.com}
}
\begin{document}
\maketitle

\input{sections/00_abstract}
\input{sections/01_intro_related_work}

\input{sections/02_pipeline}

\input{sections/04_evaluation}

\input{sections/05_results}
\input{sections/07_conclusion}
\input{sections/06_limitations}

% \clearpage
\bibliography{custom}

\clearpage 
\appendix % Supplementary figure and table numbering 
\setcounter{figure}{0} \setcounter{table}{0} \renewcommand{\thefigure}{S.\arabic{figure}} \renewcommand{\thetable}{S.\arabic{table}} % Unique internal identifiers for hyperlinks 
\renewcommand{\theHfigure}{supp.\arabic{figure}} \renewcommand{\theHtable}{supp.\arabic{table}}
\input{sections/appendix}

\end{document}

%% file: sections/00_abstract.tex
\begin{abstract}

Cross-lingual information retrieval (CLIR) is increasingly important in multi-national industries, where critical technical evidence may exist in a different language than the query. However, existing benchmarks do not adequately capture domain-specific cross-lingual retrieval or the retrieval-depth and recoverability failures that aggregate recall hides.
In this work, we benchmark CLIR in the chemical domain, with a focus on patent data. We construct a multilingual dataset from Google Patents and the European Patent Office (EPO) data, spanning five languages (covering major Eastern and Western languages) and reflecting the diversity and complexity of real-world industrial documentation.
Using this dataset, we systematically evaluate eight state-of-the-art embedding models for cross-lingual retrieval. Our results show a substantial performance gap between monolingual and cross-lingual settings: for the best-performing model, Recall@10 drops from 0.72 to 0.53 in cross-lingual setting. Retrieval depth also degrades significantly, with relevant documents ranked lower across languages in cross-lingual scenarios. Furthermore, some multilingual embedding models that perform strongly in monolingual settings exhibit sharp declines when queries and documents are in different languages, providing practical insights for model selection in cross-lingual use cases.
These findings highlight critical limitations of current approaches and emphasize the need for more robust cross-lingual retrieval methods in domain-specific settings. Our benchmark provides actionable insights for model selection and establishes a controlled diagnostic evaluation framework for CLIR over industrial technical text. Data and code are publicly available \href{https://github.com/MohammadKhodadad/Multi-Lingual-QAC}{here}.
\end{abstract}

%% file: sections/01_intro_related_work.tex
\section{Introduction and Related Work}

% Technical chemistry search rarely respects a single language boundary: a chemist
% may query in English, German, French, or Spanish while relevant evidence resides
% in a multilingual patent publication, a regional filing, or a document whose
% language differs from the query. This is not only a translation problem:
% chemistry retrieval must also distinguish a target compound from chemically
% related but incorrect alternatives. A retriever for a high-trust workflow such as
% patent search or RAG therefore has to do two hard things at once: cross the
% language barrier and resist chemical confusability.

% Technical chemistry search rarely respects a single language boundary: a chemist
% may query in English, German, French, or Spanish while relevant evidence resides
% in a multilingual patent publication, a regional filing, or a document whose
% language differs from the query. This is not only a translation problem:
% chemistry retrieval must also distinguish a target compound from chemically
% related but incorrect alternatives. A retriever for a high-trust workflow such as
% patent search or RAG therefore has to do two hard things at once: cross the
% language barrier and resist chemical confusability.

Cross-lingual information retrieval (CLIR) is increasingly critical in multi-national industrial environments, where proprietary technical data is distributed across multiple languages. In domains such as chemistry, relevant evidence for a given query may exist only in a different language, and retrieval failures can lead to missed prior art, incomplete regulatory analysis, or suboptimal technical decisions. These challenges are further amplified in agentic AI systems and retrieval-augmented generation (RAG) pipelines, where retrieval quality directly impacts downstream answer generation and reasoning \citep{kukreja2026study}.

Machine translation is commonly used to bridge language gaps. However, translation-based retrieval introduces additional computational cost and latency, and may not reliably preserve specialized terminology in technical domains. \citep{valentini2025clirudit,kim2024efficient}. Recent advances in dense retrieval using multilingual embeddings \citep{wang2024multilingual,labse2022}
enable unified indexing across languages, but robust cross-lingual alignment remains a core challenge \citep{goworek2025bridging}. Prior work shows that retrieval behavior is strongly influenced by alignment quality, translation effects, and same-language preference \citep{whatdrivesclir2025,crosslingualcost2025,bordirlines2024,xrag2025,nepotism2025}. Consequently, aggregate metrics such as Recall@$k$ may mask cross-lingual failures.

Existing benchmarks address related aspects of this problem. Multilingual retrieval benchmarks such as MMTEB, MIRACL, NeuCLIR, and CLIRMatrix provide broad multi-lingual and cross-language evaluation \citep{miracl2023,neuclir2023,mmteb2025,clirmatrix2020}. In parallel, chemistry-focused resources evaluate domain-specific capabilities in predominantly monolingual settings, including text embedding and retrieval with ChemTEB and the ChemRxiv Retrieval benchmark, RAG with ChemLit-QA, and compositional or multi-hop reasoning with ChemComp and ChemKGMultiHopQA \citep{chemteb2024,chemlitqa2024,khodadad2026chemcomp,astaraki2026iterativerag,chembed2025}. However, these lines of work remain disconnected: multilingual benchmarks lack domain-specific technical grounding, and chemistry benchmarks do not evaluate cross-lingual retrieval. Appendix~\ref{app:benchmarks} tabulates this comparison. Moreover, prior work does not systematically distinguish same-language and cross-language retrieval or capture deployment-relevant behaviours such as retrieval depth and recoverability.

To address these gaps, we introduce ChemCLIR-Bench, a benchmark for cross-lingual retrieval framed as patent-grounded parallel-document retrieval: a query must recover its source chemical patent in the query's language and in the patent's other language versions.. We construct two multilingual question–answer–context (QAC) datasets derived from Google Patents and the European Patent Office (EPO), spanning five languages. Patents provide naturally aligned, high-quality multilingual technical content, enabling controlled evaluation in realistic industrial settings.
Our evaluation framework is designed to surface deployment-relevant failures by explicitly separating monolingual (MLIR) and cross-lingual (CLIR) retrieval, and by introducing diagnostic metrics that capture retrieval depth, reading cost, and re-ranking recoverability.
We evaluate eight multilingual embedding models and show that cross-lingual retrieval remains a major bottleneck: all models exhibit a consistent drop in performance relative to monolingual retrieval, and many failures cannot be recovered by downstream re-ranking.
Our contributions can be summarized as: (i) A multilingual benchmark for patent-grounded parallel-document retrieval over chemical patents.
(ii) (ii) We propose a language-aware evaluation framework that goes beyond Recall@k and captures retrieval-depth and recoverability behaviour.
(iii) We provide a systematic analysis showing that cross-lingual alignment, rather than overall retrieval capacity, is the primary limitation in current models.

%% file: sections/02_pipeline.tex
\section{QAC Generation Pipeline}

\begin{figure*}[t]
    \centering
    \includegraphics[width=0.98\textwidth]{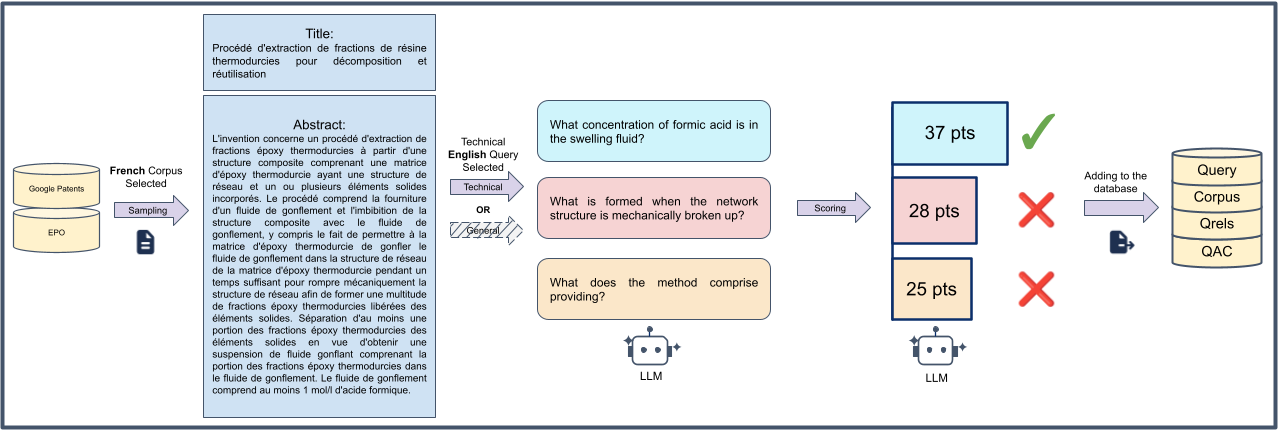}
    \caption{Overview of the QAC generation pipeline: source documents are
    converted into grounded candidate questions, scored, and exported as
    retrieval-ready records. The pipeline optimized with three rounds of domain expert feedback collection. %Icons adapted from Font Awesome, licensed under CC BY 4.0. 
    }
    \label{fig:qac-generation-flow}
\end{figure*}

Our pipeline turns multilingual chemistry patent text into
question--answer--context (QAC) records for retrieval evaluation. It produces
questions in five languages, designed for cross-lingual retrieval and kept close
to what a practitioner would actually search, in both technical and general
modes. Figure~\ref{fig:qac-generation-flow} summarizes the generation and
evaluation flow.

\paragraph{Corpus and context construction.}
We build on two patent-derived corpora that contribute complementary kinds of
multilinguality. \textit{Google Patents} supplies scale and language breadth:
23{,}787 published applications, predominantly World Intellectual Property Office:WIPO (WO, 15{,}048) and European
(EP, 4{,}610) filings, with a sizeable Mexican (MX, 3{,}952) share, spanning five
languages (English, French, Spanish, German, Chinese), from which we take the
title and abstract as the evidence block. The English, French, Spanish, and German text comes from the language-tagged \texttt{title\_localized} and \texttt{abstract\_localized} fields of the Google Patents Public Data table, whose schema describes them as publication text in each language; we do not use the machine-translated fields of the separate Google Patents Research dataset (Appendix~\ref{app:corpus}). Because the metadata does not establish whether every localized record was originally authored in its tagged language, we describe these records as language-tagged bibliographic patent text rather than as native or as machine-translated. The only translation step we introduce is for Chinese: coverage was zero, so we machine-translated 400 documents with GPT 5.5 to seed it, and a native-speaker audit of this subset is reported in Appendix~\ref{chinese_translation}. \textit{EPO} supplies clean
cross-lingual parallelism: 11{,}315 granted B1 specifications, each published as a human-translated English/German/French triple ($\approx$3{,}770 per
language) with professionally translated first claims, which we add to the
evidence block; we introduce no translations of our own on this side.

% \paragraph{Candidate generation.}
\paragraph{Query generation.}
For each source document we prompt \texttt{gpt-5-mini} to generate three
candidate questions, each grounded completely by the passage. Table~\ref{tab:benchmark-stats} provide an overview of the dataset and its statistics. 
\begin{table*}[b]
\centering
\footnotesize
\begin{tabular*}{\textwidth}{@{\extracolsep{\fill}}lrrrrrrrr@{}}
  \toprule
  Source & Corpus & Queries & Qrels & Cross-lang qrels & Q tokens & Q vocab & C tokens & C vocab \\
  \midrule
  Google Patents & 23,787 & 524 & 1,284 & 1,023 & 11.3 $\pm$ 6.3 & 2,742 & 145.0 $\pm$ 63.8 & 83,333 \\
  EPO & 11,315 & 198 & 594 & 396 & 14.1 $\pm$ 5.4 & 1,371 & 178.2 $\pm$ 71.1 & 72,256 \\
  \bottomrule
\end{tabular*}
\caption{Statistics of the two released benchmarks. A corpus document (C) is a title plus abstract (Google Patents) or title, abstract, and first claim (EPO) in one language; Q denotes queries. Token columns give the mean $\pm$ standard deviation per query or per document; vocabulary columns give the number of distinct case-folded tokens over all queries or all documents of a benchmark, pooled across languages.}
\label{tab:benchmark-stats}
\end{table*}
% Two choices shape the query distribution. First, a question's \emph{target language} is sampled per document relative to the languages that document exists in: \emph{random-existing} picks a language the patent already has a version in, so the query has a same-language answer; \emph{random-missing} picks one it does \emph{not}, yielding a pure cross-lingual ``no-home'' query; and \emph{all} generates the question in every 5 languages we targeted. Second, each question is written in one of two \emph{modes}: \emph{technical} questions ask for a single concrete fact, a parameter, material, outcome, method, or structural detail, answerable by the document, while \emph{general} questions ask about the document's problem, approach, or application in deliberately paraphrased, low-overlap wording. The mode prompts are written in the target language so questions read natively. The prompts (English version) are available in  Figures~\ref{fig:prompt-technical} and~\ref{fig:prompt-general}.
Two choices shape the query distribution. First, the query \emph{language}: the queries are generated in three set ups, where the question's
\emph{target language} is sampled per document relative to the languages that
document exists in: (i) \emph{random-existing}: picks a language the patent already
has a version in, so the query has both a cross-lingual gold context and a same-language gold context; (ii)
\emph{random-missing}: picks a language it does \emph{not} have a document in, yielding a pure cross-lingual
``no-home'' query, where no same-language gold-context is available for the query; and (iii) \emph{all} generates the question in every 5 languages we targeted, independent of availability of the gold-context document in those languages. Second, each question is written in one of two \emph{modes}: (i)
\emph{technical} questions ask for a single concrete fact, a parameter, a
material, an outcome, method, or structural detail, answerable by the document, while (ii) \emph{general} questions ask about the document's problem, approach,
or application in deliberately paraphrased, low-overlap wording. The mode prompts
are written in the target language to bring the generated questions closer to native quality. The prompts (English version) are available in  Figures~\ref{fig:prompt-technical}
and~\ref{fig:prompt-general}.

The three generated candidates per document for each query are scored not by the generator but by a
separate verifier, \texttt{claude-sonnet-4.6}, which rates each candidate
on multiple metrics around faithfulness and mode-specific quality; we keep the highest-scoring
candidate per document as the selected query (more details on verifier prompts are available in
Figures~\ref{fig:verifier-faith}--\ref{fig:verifier-general}). We then export the standard retrieval artifacts.
In the process of query generation, the full generation pipeline benefited from three
rounds of domain-expert annotation and feedback collection which significantly refined
both the generation quality and the verifier prompts. In the final set of generated
queries, the verifier was audited against human-annotation on a 106-sample subset of the
queries using criteria that mirror the LLM rubric (see
section~\ref{ssubsec:human_annotation_query} for more details). This annotated subset
corresponds to approximately 14.7\% of all generated queries. In this evaluation, the
human feedback's mean score was \textbf{8.40/10} (technical: \textbf{8.78}, general:
\textbf{8.00}) and \textbf{no} query was rejected (none scored in the 0--3 band; see
Figure~\ref{fig:human-scores} for more details). Also, the agreement between LLM-verifier
and human expert annotator is reported in \ref{human_model_aggreement}.

% \paragraph{Domain Expert Human Evaluation}
% The human evaluation was conducted on 10\% of the queries, focusing on faithfulness, retrieval-oriented query quality, and linguistic quality. All evaluation criteria were defined in alignment with prior work in the literature. Details of the criteria, a sample screenshot of the annotation tool, and qualitative feedback from annotators are provided in the supplementary materials~\ref{ssubsec:human_annotatio}. The raw collected feedback is open-sourced and available at \href{?}{here\textcolor{red}{reference to be added}}.

%% file: sections/04_evaluation.tex
\section{Evaluation Setup}
\label{sec:setup}

\begin{table*}[h!]
\centering
\small
\setlength{\tabcolsep}{4.5pt}
\begin{tabular}{l c cc cc ccc}
\toprule
 & & \multicolumn{2}{c}{\textbf{Google Patents}} & \multicolumn{2}{c}{\textbf{EPO}} & \multicolumn{3}{c}{\textbf{Optimum K for retrieval}} \\
\cmidrule(lr){3-4}\cmidrule(lr){5-6}\cmidrule(lr){7-9}
Model & Size & R@10$_{\mathrm{same}}$ & R@10$_{\mathrm{cross}}$ & R@10$_{\mathrm{same}}$ & R@10$_{\mathrm{cross}}$ & $k^\star_{80}$ & $k_{\mathrm{80 same}}$ & $k_{\mathrm{80 cross}}$ \\
\midrule
embeddinggemma            & 300M & \textbf{0.74} & \textbf{0.54} & 0.70          & \textbf{0.52} & \textbf{147}\,$\pm$\,38 & \textbf{36}\,$\pm$\,13 & \textbf{121}\,$\pm$\,40 \\
bge-m3                    & 568M & 0.64          & 0.47          & \textbf{0.71} & 0.47          & 304\,$\pm$\,75          & 66\,$\pm$\,24          & 215\,$\pm$\,45 \\
qwen3                     & 600M & 0.62          & 0.45          & 0.63          & 0.39          & 425\,$\pm$\,93          & 63\,$\pm$\,16          & 310\,$\pm$\,82 \\
nomic-v2-moe              & 475M & 0.65          & 0.41          & 0.67          & 0.43          & 367\,$\pm$\,77          & 55\,$\pm$\,14          & 321\,$\pm$\,68 \\
granite                   & 278M & 0.54          & 0.39          & 0.43          & 0.36          & $>$1000                 & 641\,$\pm$\,167        & 569\,$\pm$\,153 \\
LaBSE                     & 471M & 0.45          & 0.27          & 0.30          & 0.25          & $>$1000                 & 761\,$\pm$\,136        & $>$1000 \\
e5-large-instruct$^\dagger$ & 560M & 0.73        & 0.09          & 0.63          & 0.12          & $>$1000                 & 47\,$\pm$\,14          & $>$1000 \\
SapBERT                   & 270M & 0.40          & 0.20          & 0.31          & 0.17          & $>$1000                 & 807\,$\pm$\,108        & $>$1000 \\
\bottomrule
\end{tabular}
\caption{Retrieval performance (Recall@10) on both patent datasets, split by language route: R@10$_{\mathrm{same}}$ is same-language (monolingual) and R@10$_{\mathrm{cross}}$ is cross-language, each language-balanced. The $k$ columns are right-censored 80th-percentile retrieval depths (lower is better; $\pm$ values are bootstrap standard errors over queries, 2000 resamples; omitted where the depth is censored at $>$1000). Best per column in bold. Size is total parameters.}
\label{tab:main-results}
\end{table*}

We evaluate multilingual embedding models as drop-in dense retrievers, and design
the setup to surface cross-lingual failure rather than average it away. Each query
is judged against its source patent in both its own language and every
other, separating same-language from cross-language relevance. We report a
metric family that places capability, cross-linguality, and deployment cost side
by side, and run it over a fixed set of eight models against the shared
multilingual corpus.

\paragraph{Same- vs.\ cross-language relevance.}
A query is relevant to its source patent in every language that patent
exists in, so we split each judgment into same-language relevance and cross-language relevance
(a foreign-language version to the query). Some
queries have no same-language gold at all, a built-in ``no-home'' cross-lingual stress test. We retain query
language, document language, source office, and question mode (technical or
general) as metadata, so every result can be sliced along each axis.

% \paragraph{Alias-graph stress test.}
% The alias graph asks a sharper question: given a concept named in several
% languages, can a retriever find the right document across languages while
% ignoring documents about chemically similar concepts? We build it from the ChEBI
% ontology \citep{chebi2016}, a structured hierarchical representation of chemical compounds, maintained by European Bioinformatics Institute. For each compound we take a source patent passage
% that mentions the target compound and prompt the generator with a technical-style prompt, but with
% the compound supplied, to write a question grounded in that passage whose answer
% is the compound itself, never naming the compound or any of its multilingual
% aliases (150 questions generated). The gold context for retrieval is then the source publication and its parallel language versions, so a query in one language must reach that publication in another;
% the hard negatives are patents that mention a taxonomic neighbour of the compound
% (a sibling or parent in the graph) but not the compound itself: chemically close,
% but the wrong answer. These near-miss distractors operationalize, at the
% retrieval level, the distractor-latch failure that is destructive downstream,
% where a generator locks onto a plausible but wrong neighbour rather than the
% intended evidence \citep{astaraki2026iterativerag}.

\paragraph{Metrics.}
We report Recall@10 split by language route. A query is relevant to its source
patent in every language that patent exists in, so we separate same-language from
cross-language gold: R@10$_{\mathrm{same}}$ is Recall@10 against the query's-language
copy of the patent, and R@10$_{\mathrm{cross}}$ is Recall@10 against a copy in any
other language, each computed over the queries that have the corresponding gold type
and balanced over query languages.

The coverage-depth metrics invert Recall@10: instead of fixing the cutoff at $10$ and
measuring how much gold is found, they fix the target at $80\%$ of queries and measure
how deep one must read to reach it. They are computed over the queries that have both a
same- and a cross-language gold, pooled across both benchmarks. For each
query, let the same-language depth be the rank of the first (best-ranked) same-language
gold in the top-$1000$ retrieval list, and the cross-language depth the rank of the
first cross-language gold; a depth is infinite when no such gold appears in the top
$1000$. Then $k_{\mathrm{80 same}}$ is the smallest depth $k$ at which at least one
same-language gold falls within the top-$k$ for $80\%$ of queries (the nearest-rank
$80$th percentile of the same-language depth), $k_{\mathrm{80 cross}}$ is the same quantity
for the cross-language depth, and $k^\star_{80}$ is the smallest $k$ at which
\textbf{both} golds are within the top-$k$ for $80\%$ of queries (the $80$th percentile
of the deeper of the two depths per query). Lower is better, and all three are
right-censored at $>$1000 when fewer than $80\%$ of queries reach the target within the
retrieved top-$1000$.

To characterize how expensive it is to reach that
evidence and how much of it a downstream re-ranker could still recover.\textit{(i) Cross-lingual reading-cost multiplier (XRC).}
For a model $m$ and a coverage fraction $C\in(0,1]$, let $D^{\text{same}}_C(m)$ and
$D^{\text{cross}}_C(m)$ be the retrieval depths at which the same-language gold and
the first cross-language gold, respectively, are reached for a fraction $C$ of
queries. The reading-cost multiplier is their ratio:
\begin{equation}
\mathrm{XRC}_C(m)=\frac{D^{\text{cross}}_C(m)}{D^{\text{same}}_C(m)}.
\end{equation}
It is the factor by which a user must read deeper in the ranked list to reach a
foreign-language version of the answer than to reach a same-language copy
($\mathrm{XRC}=1$ means no penalty). The depth-to-reach quantity $D_C$ is inspired by \citet{cooper1968esl}.

\textit{(ii) Re-ranker recoverability ceiling (RRC).}
Because a re-ranker can only reorder the candidate pool that first-stage retrieval
hands it \citep{nogueira2019passage,lin2021pretrained}, the fraction of cross-lingual
queries whose first cross-language gold (at rank $r^{\text{cross}}$) appears within
the top-$K$ upper-bounds what any top-$K$ re-ranker can recover:
\begin{equation}
\mathrm{RRC@}K(m)=\Pr\big[\,r^{\text{cross}}\le K\,\big].
\end{equation}
$\mathrm{RRC@}K$ is the cumulative hit rate of the first cross-language gold within the
top-$K$, a corpus-level generalization of mate-retrieval accuracy
\citep{artetxe2019massively}. Its knee $K^\star$ marks where returns diminish, the
smallest pool capturing most recoverable cross-language gold.

\textit{(iii) Alignment-recoverability index (ARI).}
Building on RRC, we partition the cross-lingual outcome at an operating depth $K$ to
attribute the residual gap to its source. The unit mass of cross-lingual queries
splits exhaustively into three terms that sum to one: those whose foreign gold is
already in the top-$K$ ($\mathrm{RRC@}K$), those reachable only by enlarging the pool
to the full retrieved depth ($\mathrm{RRC@}1000-\mathrm{RRC@}K$), and an irreducible
floor $L_\infty=1-\mathrm{RRC@}1000$ that no re-ranker over the retrieved top-1000 can
move. ARI reports the fraction of the depth-$K$ shortfall attributable to that floor:
\begin{equation}
\begin{split}
\mathrm{ARI@}K(m)
&=\frac{L_\infty(m)}{1-\mathrm{RRC@}K(m)}\\[2pt]
&=\frac{1-\mathrm{RRC@}1000(m)}{1-\mathrm{RRC@}K(m)}.
\end{split}
\end{equation}

\paragraph{Models and protocol.}
We evaluate 8 multilingual or domain-relevant embedding models under 1 billion parameters:
\texttt{embeddinggemma}, \texttt{bge-m3} \citep{bgem3_2024},
\texttt{granite-278m}, \texttt{nomic-v2-moe}, \texttt{qwen3-0.6B},
\texttt{LaBSE} \citep{labse2022}, \texttt{SapBERT} \citep{sapbert2021},
\texttt{e5-large-instruct}. All models retrieve against
the same multilingual chemistry corpus, and all queries are evaluated with the
same relevance judgments.

%% file: sections/05_results.tex
\section{Results}
\begin{figure*}[t]
  \centering
  \includegraphics[width=.96\linewidth]{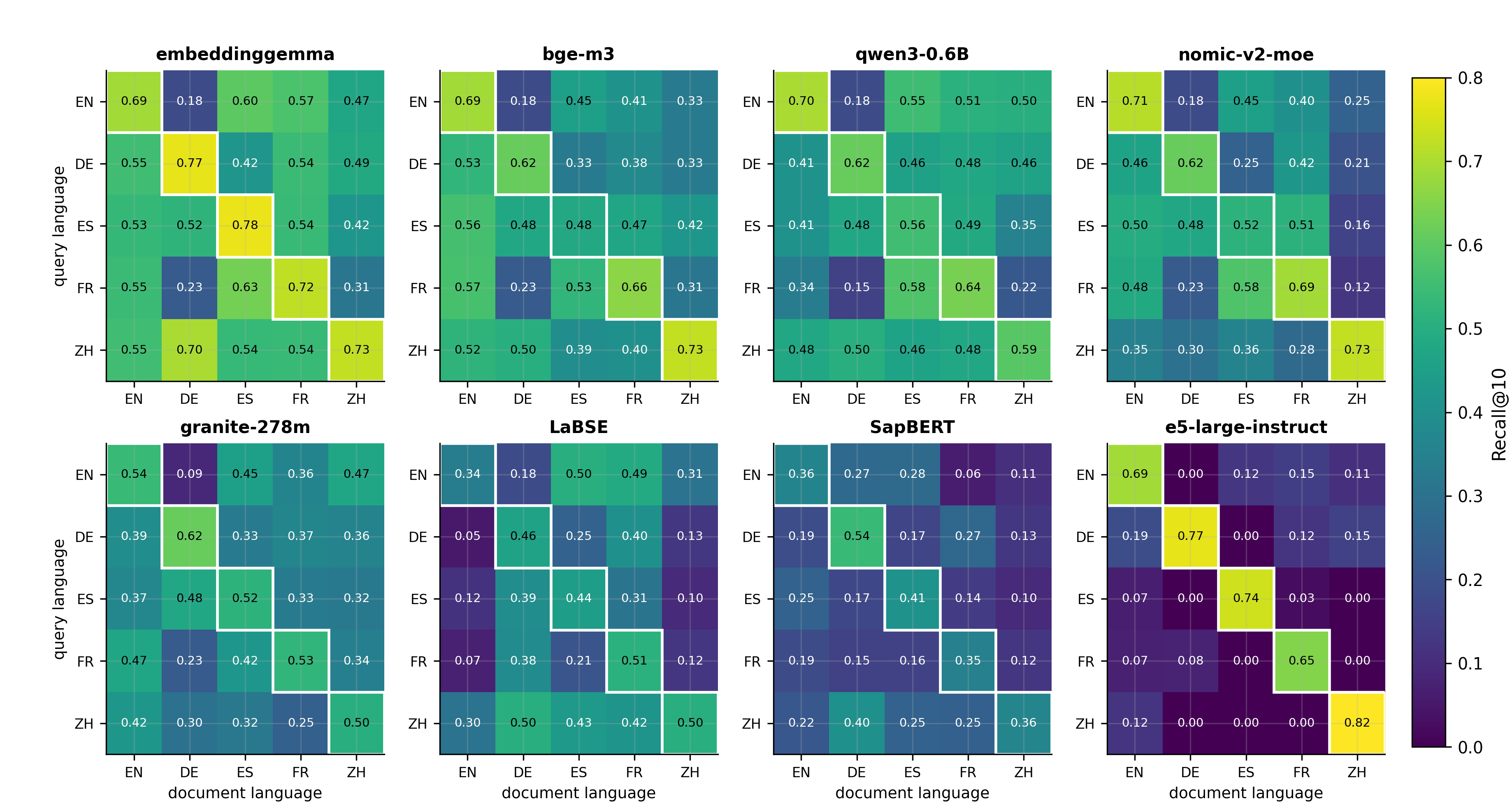}
  \caption{Recall@10 by query language $\times$ document language, per model (diagonal  = same-language; off-diagonal = cross-lingual). }
  \label{fig:qd-grid}
\end{figure*}

The performance of all evaluated models across the Google Patents and EPO corpora, averaged over the covered languages, is summarized in Table~\ref{tab:main-results}.

\subsection{Model Rankings and the Cross-Lingual Gap}
Table~\ref{tab:main-results} reports Recall@10 split by language route. Every recall
figure is language-balanced: for each model and language route we compute recall
within each of the five query languages (English, German, French, Spanish, Chinese)
and average those five values, so each query language is weighted equally rather than
letting the most frequent one dominate a pooled average. embeddinggemma leads
cross-language retrieval on both offices (R@10$_{\mathrm{cross}}$ 0.54 / 0.52) and tops
same-language recall on Google Patents (0.74); bge-m3 is a consistent second and
narrowly takes same-language EPO (0.71 vs.\ 0.70). qwen3 and nomic-v2-moe are
mid-pack, while granite, LaBSE, and SapBERT trail by 15--35 Recall@10 points. The
ordering is nearly identical across Google Patents and EPO, so it reflects the task
rather than a single corpus. One model is an instructive outlier: e5-large-instruct is
near the top on same-language recall (0.73 / 0.63, second only to embeddinggemma on
Google Patents) yet its cross-language recall collapses to 0.09 / 0.12, so a model
that looks strong under same-language recall alone fails outright once cross-language
evidence is required. Off-the-shelf multilingual embedders are therefore far from
interchangeable here, and even the best cross-language number (0.54) leaves nearly
half of the foreign-language evidence outside the top-10, low for an industrial
deployment. Deeper cutoffs raise recall without closing the gap
(Table~\ref{tab:recall-depth}): embeddinggemma's cross-language recall climbs from
0.54 at R@10 to 0.68 at R@50, but its same-language recall rises in step (0.74 to
0.85), so a same/cross gap of 0.14--0.17 persists even at depth 50, where the best
cross-language recall still misses about a third of the foreign evidence.

Across the field, retrieving a same-language copy of the answer is far easier than
retrieving its foreign twin, and no model escapes the drop. The coverage depths make
the gap concrete: among the models that retrieve effectively, the same-language gold
surfaces within a few dozen documents ($k_{\mathrm{same}}$ from $36{\pm}13$ to
$66{\pm}24$) while its cross-language twin takes several hundred ($k_{\mathrm{cross}}$
from $121{\pm}40$ to $321{\pm}68$); embeddinggemma reaches both golds the fastest,
needing $k^\star_{80}{=}147{\pm}38$ documents to cover $80\%$ of queries against
$304{\pm}75$ for the next model. These separations lie well outside the bootstrap
standard errors, so the same/cross depth gap is not a sampling artifact. For the
weaker models both depths balloon and the cross-language gold is frequently never
covered for $80\%$ of queries within the retrieved top-1000. The same pattern holds language by
language: pooled across the non-degenerate models, same-language recall stays near
0.56--0.62 in every query language while cross-language recall sits around 0.34--0.38,
a roughly 0.25 gap present in all five languages (Figure~\ref{fig:home-advantage});
and the query$\times$document grid shows the loss lives off the diagonal for every
model (Figure~\ref{fig:qd-grid}): even the strongest retrievers fall to a fraction of
their same-language cell on individual routes, embeddinggemma drops from 0.69
same-language (English) to 0.18 when an English query must reach a German document,
bge-m3 from 0.62 (German) to 0.33 reaching a Spanish document, and nomic-v2-moe from
0.69 (French) to 0.12 reaching a Chinese one. Cross-linguality, not raw retrieval
power, is what every model struggles with.

\paragraph{Uncertainty, significance, and route counts.} To quantify these differences we computed 95\% confidence intervals for R@10$_{\text{same}}$ and R@10$_{\text{cross}}$ from a language-stratified, family-clustered BCa bootstrap with 10,000 resamples; interval half-widths range from $\pm$0.02 to $\pm$0.08 recall points, so all but the smallest same-versus-cross gaps in Table~\ref{tab:main-results} lie well outside sampling error. We test the gap with paired analyses on dual-gold queries, those with both gold types ($n$=261 for Google Patents, $n$=198 for EPO), using cluster-bootstrap intervals for the paired difference $\Delta$ and family-level sign-flip permutation tests \citep{smucker2007comparison}. The gap is significant for every model on Google Patents and for all EPO models except LaBSE ($\Delta$ = +0.046, 95\% CI [$-$0.006, +0.099]). A Friedman test on per-query R@10 (same- and cross-language gold pooled) rejects equal performance across the eight models on both corpora ($\chi^2$=809 for Google Patents and 310 for EPO, $\mathrm{df}$=7, both $p<10^{-60}$), and Holm-corrected pairwise comparisons put embeddinggemma significantly ahead of all seven other models on Google Patents (all Holm-adjusted $p \le 0.0007$, all CIs excluding zero) while leaving adjacent mid-table pairs indistinguishable (bge-m3 vs.\ qwen3-0.6B, difference +0.017, 95\% CI [$-$0.009, +0.042]). Per-route counts are strongly imbalanced: the 524 Google Patents queries yield 261 same-language and 1,023 cross-language gold pairs, with per-route $n$ from 10 (zh$\rightarrow$de) to 119 (en$\rightarrow$en), which is why every headline number is a language-balanced macro mean rather than a pooled one; EPO is complete by construction, with 72/58/68 queries in en/de/fr and every route present.

\subsection{Reading Cost and the Re-Ranking Ceiling}

\begin{figure*}[t]
  \centering
  \includegraphics[width=.99\linewidth]{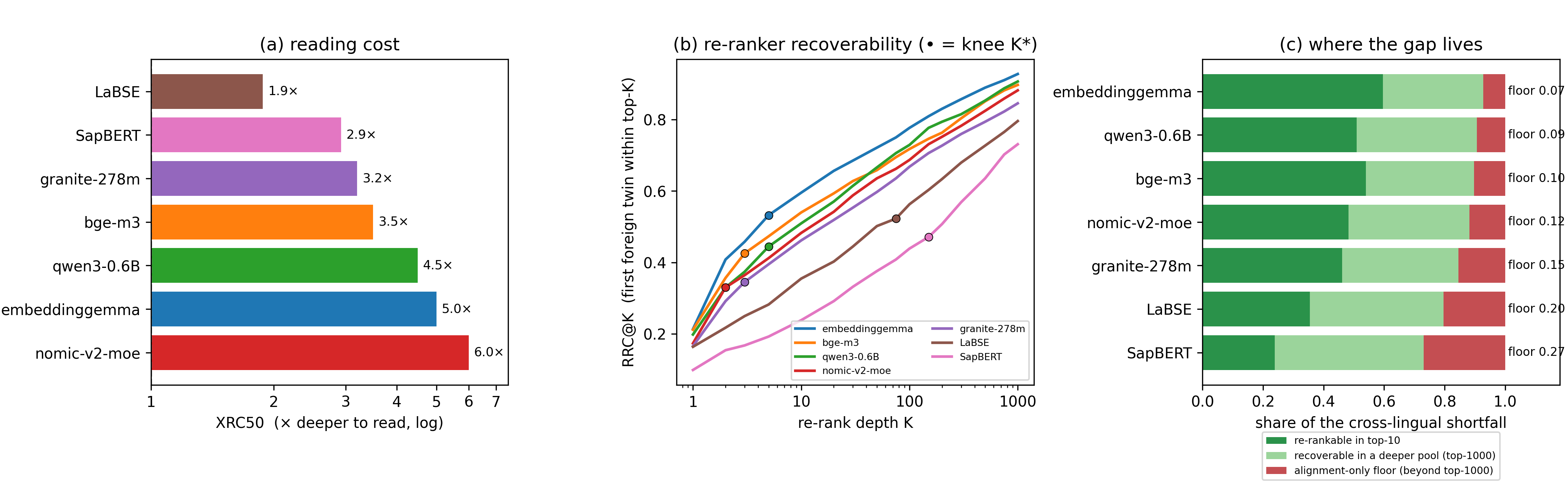}
  \caption{Deployment cost of cross-lingual retrieval (e5-large-instruct excluded by
  the R@10$_{\mathrm{cross}}<0.10$ gate). \textbf{(a)}~XRC50 reading cost; \textbf{(b)}~RRC@$K$
  re-ranker recoverability with knee $K^\star$ ($\bullet$); \textbf{(c)}~the shortfall
  split into re-rankable (top-10), deeper-pool (top-1000), and an alignment-only floor.}
  \label{fig:cost}
\end{figure*}

Panel~(a) of Figure~\ref{fig:cost} reports the median reading-cost multiplier
$\mathrm{XRC}_{50}$, and it is never one: the cheapest model still reads about twice as
deep (LaBSE, 1.9$\times$) and the rest three to six times deeper (bge-m3 3.5$\times$,
embeddinggemma 5.0$\times$, nomic-v2-moe 6.0$\times$). A low reading cost is not a
virtue on its own: the cheapest readers here are the weakest retrievers, which read
shallow only because they surface so little, so $\mathrm{XRC}$ must be read together
with capability (R@10$_{\mathrm{cross}}$) rather than minimized in isolation. We drop
e5-large-instruct from all three panels because it fails the degeneracy gate
(R@10$_{\mathrm{cross}}<0.10$): it almost never retrieves a foreign twin, so its
reading-depth ratio explodes (median 161$\times$) and its recoverability curve rests
on mostly censored ranks, the instruments are undefined for a model that does not
retrieve the thing whose depth they measure.

Panels~(b) and~(c) ask what a downstream re-ranker could do about the buried
cross-language documents. Two things stand out. First, the recoverable gold is
front-loaded: the knee $K^\star$ sits at $2$--$5$ for the strong models (Panel~b), so a
shallow top-5 pool already captures the bulk of the cheaply recoverable cross-language
gold (embeddinggemma $\mathrm{RRC@}5{=}0.53$); a deeper pool keeps helping but with
falling efficiency per decade ($\mathrm{RRC@}100{=}0.78$, $\mathrm{RRC@}1000{=}0.93$),
while the weak LaBSE and SapBERT reach their knee only at depth $75$--$150$. Second,
even an exhaustive re-ranker over the full retrieved pool hits a wall (Panel~c). Some
cross-language gold never appears in the top-1000 at all, so no re-ranker can reach it.
This alignment-only floor $L_\infty$ is small for the strong models (0.07 for
embeddinggemma) but large for the weak ones (0.27 for SapBERT): the weaker the model,
the more of its cross-lingual failure is structural rather than a matter of pool depth.
Re-ranking cannot close this part of the gap,even a top-100 re-ranker is capped at
$\mathrm{RRC@}100$ (0.78 for the best model), since it only reorders what first-stage
retrieval already surfaced. These results suggest that improvements to cross-lingual representation alignment may be needed in addition to query-time re-ranking. For more details on the introduced metrics, see Figure~\ref{fig:reading-cost}.

% \begin{figure*}[t]
%   \centering
%   \includegraphics[width=\linewidth]{figures/claimE_E2_triptych.png}
%   \caption{Deployment cost of cross-lingual retrieval (e5-large-instruct excluded by
%   the CLIR@10 $<0.10$ gate). \textbf{(a)}~XRC50 reading cost; \textbf{(b)}~RRC@$K$
%   re-ranker recoverability with knee $K^\star$ ($\bullet$); \textbf{(c)}~the shortfall
%   split into re-rankable (top-10), deeper-pool (top-1000), and an alignment-only floor.}
%   \label{fig:cost}
% \end{figure*}

%% file: sections/07_conclusion.tex
\section{Conclusion}

This work introduced ChemCLIR-Bench, a patent-grounded benchmark for cross-lingual information retrieval in the chemical domain, combining multilingual QAC datasets with language-aware diagnostic metrics.
Our evaluation of eight multilingual embedding models reveals a consistent gap between monolingual and cross-lingual retrieval: even the strongest models fail to retrieve a significant portion of cross-language evidence at practical depths. We show that standard metrics such as Recall@k can obscure these failures, while cross-lingual retrieval incurs higher reading cost and limited recoverability through re-ranking.
These results highlight that cross-lingual alignment, rather than overall retrieval capacity, is the primary bottleneck in domain-specific settings, and cannot be addressed by re-ranking alone.
Beyond benchmarking, we introduce evaluation criteria for comparing model behaviour under realistic retrieval constraints and provide a foundation for future research. While ChemCLIR-Bench focuses on controlled multilingual patent evidence and therefore does not establish transfer to all technical or general-domain retrieval settings, it provides a reusable framework for diagnosing cross-lingual retrieval failures under realistic constraints. Extending this benchmark to broader technical domains, additional languages, and retrieval approaches that combine dense, sparse, translation-based, and hybrid methods %stronger retrieval settings
is a key direction for future multilingual retrieval evaluation.

%% file: sections/06_limitations.tex
\section*{Limitations}
The suite is intentionally specialized: chemistry patents give controlled multilingual technical evidence, but the results, and the model ordering, may not transfer to other domains, regulatory corpora, or general web retrieval. Although patent variants are
content-controlled, they are not guaranteed to be claim-level equivalent across every language. Accordingly, ChemCLIR-Bench is best viewed as a reusable evaluation framework rather than a universal assessment of multilingual retrieval performance. The QAC data rests on LLM generation and LLM-based verification; our
97-item human audit supports its quality but is limited and does not replace comprehensive expert review of all benchmark instances.
%per-item expert chemistry or patent review.%, which we leave to future work (especially for
% borderline relevance and hard-negative severity). Finally, we report only the two QAC
% benchmarks and the alias-graph diagnostics; code-switching and noisy-query settings
% exist in the project but are not claimed here.

%% file: sections/appendix.tex
\section{Appendix}
\subsection{Comparison with Related Benchmarks}
\label{app:benchmarks}
\begin{table*}[t]
\centering
\footnotesize
\setlength{\tabcolsep}{3pt}
\begin{tabular}{@{}lllllll@{}}
\toprule
\textbf{Benchmark} & \textbf{Domain} & \textbf{Lang.} & \textbf{Setting} & \textbf{Corpus} & \textbf{Relevance} & \textbf{Objective} \\
\midrule
\textbf{ChemCLIR-Bench} (ours) & Chem.\ patents & 5 & Mono + CL & Patents (GP, EPO) & Parallel-doc & CLIR diagnostics \\
MIRACL & General & 18 & Mono & Wikipedia & Human & Retrieval \\
MMTEB & General & 250+ & Mono (some CL) & Aggregated & Inherited & Emb.\ suite \\
CLIRMatrix & General & 139 / 8 & CL & Wikipedia & Synthetic & Retrieval \\
NeuCLIR & News (+tech.) & EN$\rightarrow$3 & CL, MLIR & News, abstracts & Human & Retrieval \\
ChemTEB & Chemistry & 1 (EN) & Mono & PubChem, Wiki & DB-derived & Emb.\ suite \\
ChemLit-QA & Chem.\ papers & 1 (EN) & Mono & QAC triplets & LLM + expert & RAG QA \\
\bottomrule
\end{tabular}
\caption{ChemCLIR-Bench alongside general multilingual IR benchmarks and chemistry-focused benchmarks. Column values are abbreviated; Appendix~\ref{app:benchmarks} defines them and gives the details behind each cell.}
\label{tab:benchmark-comparison}
\end{table*}

Table~\ref{tab:benchmark-comparison} positions ChemCLIR-Bench against the multilingual retrieval benchmarks and chemistry resources discussed in Section~1, with abbreviated cells. Lang.\ is the number of languages covered: CLIRMatrix lists its bilingual release (BI-139, 139 languages in 139$\times$138 pairs) and its multilingual release (MULTI-8, eight languages), and NeuCLIR uses English topics against three document languages (Chinese, Persian, Russian). Setting distinguishes monolingual retrieval within each language (Mono), cross-lingual retrieval where query and document languages differ (CL), and retrieval over a pooled multi-language collection (MLIR). ChemCLIR-Bench scores same-language and cross-language routes separately, including ``no-home'' queries with no same-language gold; MMTEB is mostly monolingual with some cross-lingual retrieval and bitext-mining tasks; ChemTEB's bitext-mining tasks pair SMILES strings with text rather than two languages, so it remains monolingual. Corpus names the document collection: patent titles and abstracts from Google Patents and EPO (EPO adds first claims) for ChemCLIR-Bench; Wikipedia passages or articles for MIRACL and CLIRMatrix, the latter using article titles as queries; more than 500 aggregated existing datasets across ten task types for MMTEB; news articles plus Chinese academic abstracts for NeuCLIR; PubChem, chemistry Wikipedia, chemistry subsets of BeIR (NQ, HotpotQA), COCONUT, and safety data sheets for ChemTEB; and 1,054 question--answer--context triplets from chemistry papers for ChemLit-QA. Relevance describes how labels were obtained. Parallel-doc means a query is relevant to every language version of its source patent, with queries LLM-generated, LLM-verified, and human-audited. Human means human-written queries with native-speaker judgments (MIRACL) or TREC-style pooled assessments (NeuCLIR). Synthetic means monolingual BM25 scores discretized into graded labels (0--6) and propagated across languages via Wikidata links. Inherited means labels taken from each constituent dataset. DB-derived means labels derived from source databases, with tasks designed or validated by chemists. LLM + expert means GPT-4 Turbo generated triplets validated by chemistry experts, with -neg (unanswerable) and -multi (multi-hop) subsets. Objective gives what the benchmark measures: ad hoc retrieval quality (Retrieval; nDCG@10 and Recall@100 for MIRACL, a re-ranking setup for CLIRMatrix, CLIR and MLIR system evaluation plus a 2024 report-generation task for NeuCLIR); multi-task embedding evaluation in which retrieval is one category among several (Emb.\ suite; ten categories for MMTEB, 34 tasks in five categories for ChemTEB); answer correctness of RAG and fine-tuned LLMs rather than retrieval ranking (RAG QA); and, for ChemCLIR-Bench, diagnosis of the cross-lingual retrieval gap and its deployment cost through R@10$_{\mathrm{same}}$ versus R@10$_{\mathrm{cross}}$, $k_{80}$ depth, XRC, RRC@$K$, and ARI.

Read together, the rows show the gap ChemCLIR-Bench fills. The general benchmarks differ mainly in how they treat language: MIRACL is monolingual within each language, MMTEB is largely monolingual with a small cross-lingual component, and CLIRMatrix and NeuCLIR are cross-lingual but built on Wikipedia and news. Only NeuCLIR includes technical text, through its Chinese academic-abstract subtask, and none is grounded in a single domain whose terminology drives relevance. The chemistry benchmarks are the mirror image: ChemTEB and ChemLit-QA are domain-specific but English-only and do not evaluate retrieval across languages. ChemCLIR-Bench sits at the intersection. It is grounded in chemical patents, covers five languages, derives relevance from a patent's own language versions rather than from propagated lexical scores, and reports same-language and cross-language recall separately together with depth-based diagnostics that the other benchmarks do not report.

\subsection{Results per query language}
\begin{figure}[h]
  \centering
  \includegraphics[width=\linewidth]{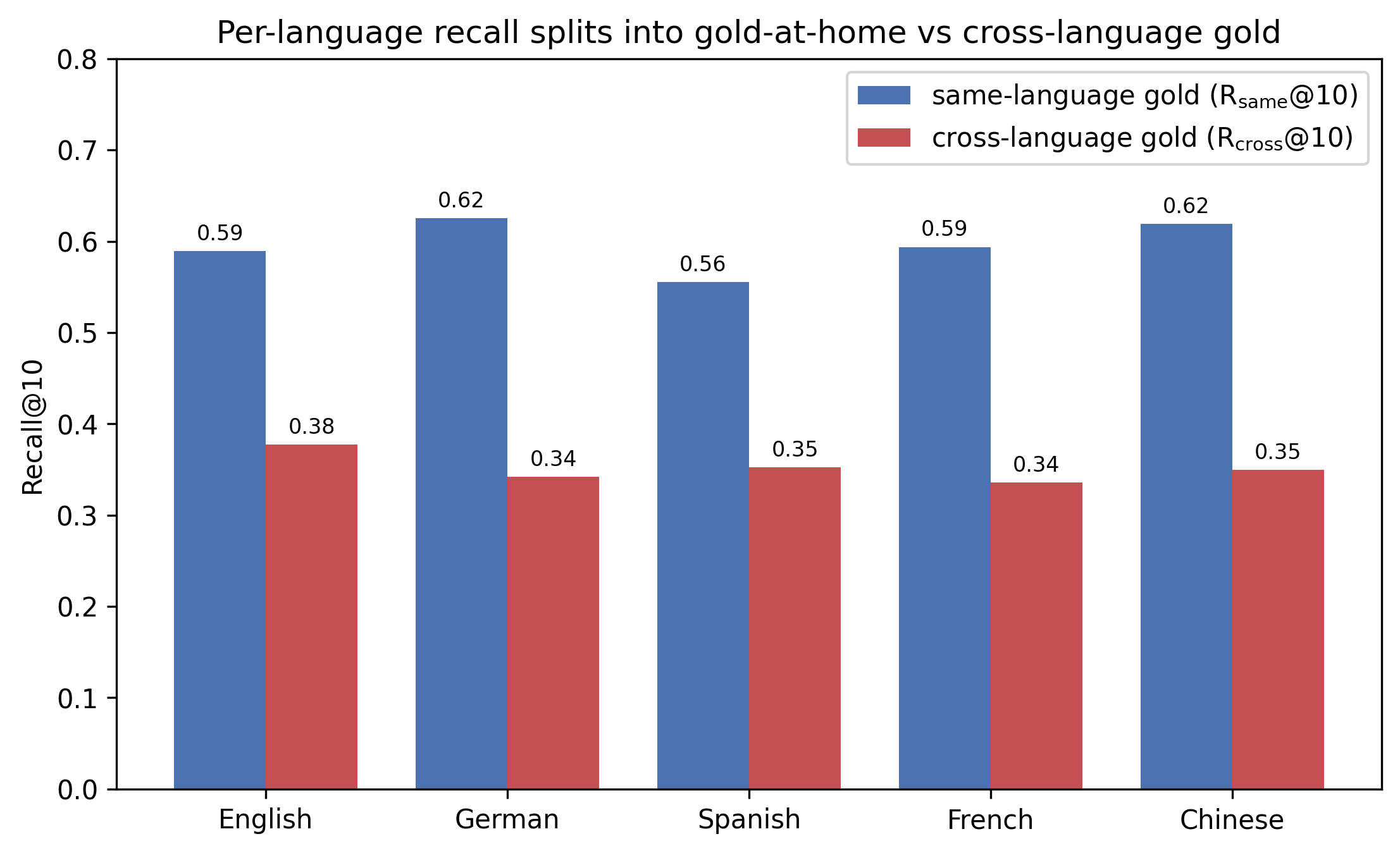}
  \caption{Per query language, Recall@10 splits into same-language gold (R@10$_{\mathrm{same}}$)
  and cross-language gold (R@10$_{\mathrm{cross}}$), pooled over the non-degenerate models. The home
  advantage holds in every language.}
  \label{fig:home-advantage}
\end{figure}

Figure~\ref{fig:home-advantage} and Figure~\ref{fig:qd-grid} reflects the break down of the performance of the models over languages. 

% \begin{figure*}[h]
%   \centering
%   \includegraphics[width=\linewidth]{figC6_query_doc_recall_grid.png}
%   \caption{Recall@10 by query language $\times$ document language, per model (white
%   cell = same-language; off-diagonal = cross-lingual). The cross-lingual loss lives
%   off the diagonal; e5-large-instruct's off-diagonal cells collapse to near zero.}
%   \label{fig:qd-grid}
% \end{figure*}

\subsection{Additional Human Annotation results}
\label{ssubsec:human_annotatio}

\subsubsection{Query quality evaluation: Human-annotation Criteria}
\label{ssubsec:human_annotation_query}
Our human evaluation rubric is organized around three families of criteria: faithfulness, retrieval-oriented query quality, and linguistic quality. The faithfulness criteria evaluate whether the generated answer is supported by the passage. First, grounding asks whether the answer can be traced to a contiguous span of the passage, following the extractive QA assumption in SQuAD, where answers are defined as spans from the source passage \citep{rajpurkar2016squad}. Second, precision asks whether the answer avoids unsupported details or padding; this adapts the notion of factual precision from FActScore, which evaluates generated text by decomposing it into atomic facts and checking what proportion is supported by a reliable source \citep{min2023factscore}. Third, numerical fidelity separately evaluates whether numbers, dates, units, and ranges are reproduced exactly. We include this as a specific subtype of factual consistency because even small numerical deviations can alter the meaning of an answer; this criterion is aligned with factual-consistency evaluation work such as QAFactEval, which checks whether generated content is supported by the source \citep{fabbri2022qafacteval}.

The query-quality criteria adapt prior QG and information-retrieval evaluation ideas to two generation modes. In technical mode, search-bar realism asks whether the generated query resembles a realistic search-engine query, motivated by fact-checking QG work where generated questions are used to retrieve evidence \citep{ousidhoum2022varifocal}. Specificity and retrievability are motivated by information-retrieval work on query clarity, which treats ambiguous queries as harder retrieval problems \citep{cronen2002predicting}. Phrasing economy extends the conciseness criterion used in QGEval and Cheng et al., while also penalizing queries that are unnecessarily copied from the passage \citep{fu2024qgeval,cheng2021guiding}. Focus asks whether the query targets a single fact, following the intuition behind answerability and answer matching: a generated question should have a clear intended answer \citep{nema2018better,cheng2021guiding,fu2024qgeval}. In general mode, search realism and retrievability again evaluate whether the query is useful for retrieval, but the emphasis shifts from keyword-style search to meaning-based retrieval. Lexical distance rewards meaningful paraphrasing rather than surface copying, following prior work emphasizing diversity in question generation \citep{sultan2020diversity}. Conceptual framing is our extension of prior QG evaluation: it asks whether the query targets a concept rather than a surface fact, inspired by work on deeper and more controllable question generation where question quality depends not only on fluency but also on the reasoning or semantic structure being elicited \citep{cheng2021guiding}. Finally, linguistic quality follows standard human evaluation practice in QG and NLG, corresponding to fluency, well-formedness, and clarity \citep{fu2024qgeval,cheng2021guiding}.

\paragraph{Human-model grading agreement.}
\label{human_model_aggreement}
To validate the LLM auto-grader against human judgment, one annotator scored a
random subset of $N{=}97$ generated questions on a $0$--$10$ quality scale, and
the auto-grader's composite score was linearly rescaled from its native
$0$--$40$ range onto the same $0$--$10$ axis. In absolute terms the two graders
are close: the mean absolute error (MAE) between the human and model scores is
$0.82$ points out of $10$. The annotated sample is near-ceiling-nearly all
questions (94 of 97 for the human, 97 of 97 for the model) fall in the top
(``good'') quality band,  so chance-corrected coefficients that assume balanced
categories are unstable: Cohen's $\kappa$ collapses to $\approx 0$ despite
$96.9\%$ raw bucket agreement, the well-known high-prevalence paradox. We
therefore report Gwet's AC1, which is robust to skewed marginals, and obtain
AC1 $=0.97$, indicating near-perfect categorical agreement. Together these
figures show that the auto-grader closely tracks human scores on this sample;
the near-ceiling distribution means the comparison reflects agreement among
predominantly high-quality questions rather than the graders' ability to flag
low-quality ones.

\begin{figure}[t]
  \centering
  \includegraphics[width=\linewidth]{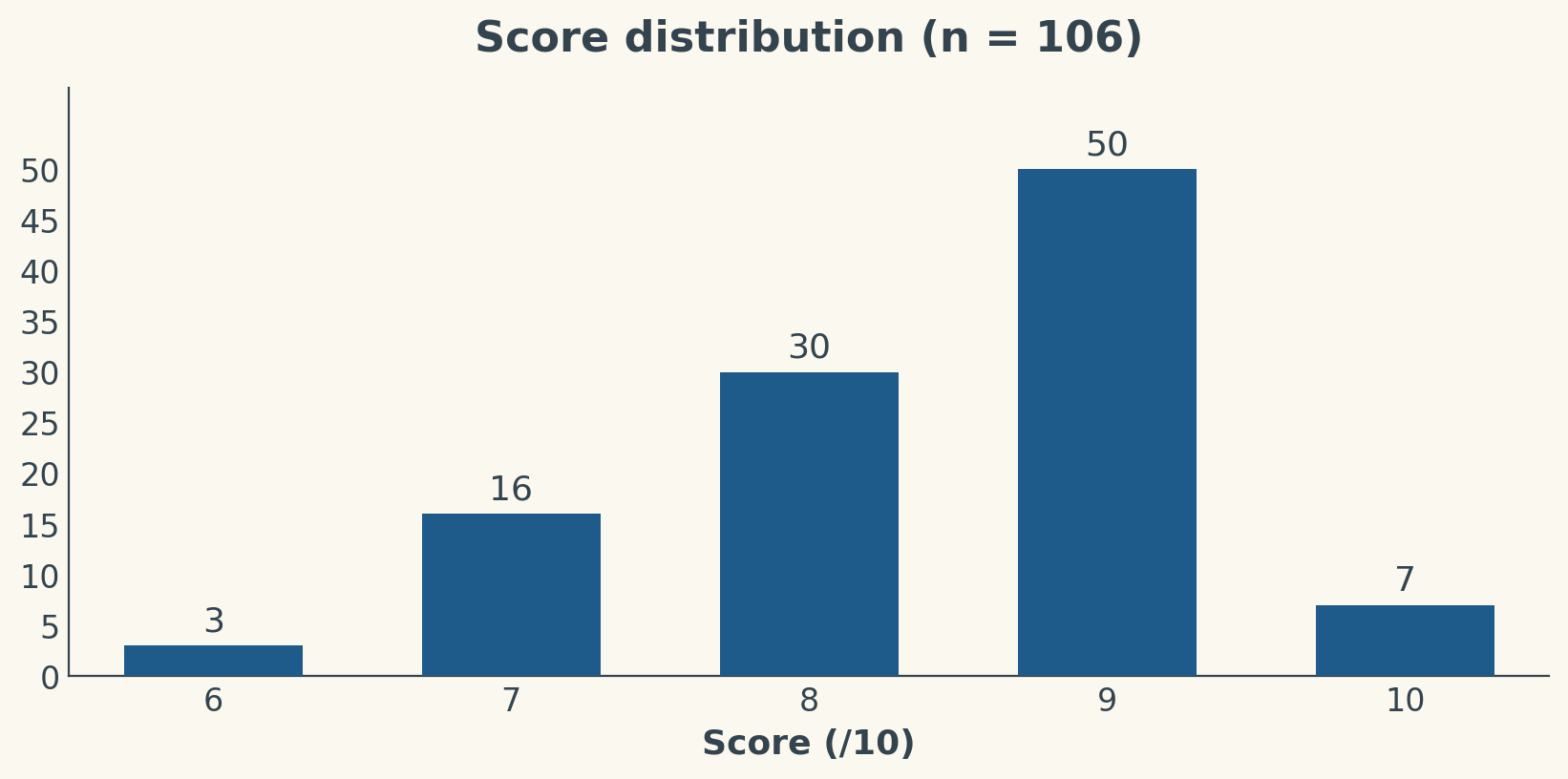}
  \caption{Distribution of human-annotator overall scores on the 106-item
  audited QAC sample (mean 8.33/10; no item rated below 6).}
  \label{fig:human-scores}
\end{figure}

\paragraph{Chinese translation Human Annotation}
\label{chinese_translation}
For the 400 Chinese abstracts we machine-translated to seed Chinese coverage, the only translation step in our pipeline, we used a manual review setup following prior MT evaluation practice, where human judges assess
adequacy and fluency \citep{koehn-monz-2006-manual}. The annotator compared the
English source with the Chinese translation, assigned 1--5 adequacy and fluency
scores, and accepted or rejected each item. Across the 40 reviewed abstracts, the
translations received a mean adequacy of $4.62/5$ and a mean fluency of $4.75/5$. Overall, the annotator accepted 34 of the 40 translations
($85\%$) and rejected only six. Figure~\ref{fig:argilla-annotation} shows the
annotation interface.

\clearpage
\onecolumn

\begin{table*}[!t]
  \centering
  \small
  \setlength{\tabcolsep}{4pt}
  \begin{tabular}{l c cc cc cc cc}
    \toprule
     & & \multicolumn{4}{c}{\textbf{Google Patents}} & \multicolumn{4}{c}{\textbf{EPO}} \\
    \cmidrule(lr){3-6}\cmidrule(lr){7-10}
     & & \multicolumn{2}{c}{R@25} & \multicolumn{2}{c}{R@50} & \multicolumn{2}{c}{R@25} & \multicolumn{2}{c}{R@50} \\
    \cmidrule(lr){3-4}\cmidrule(lr){5-6}\cmidrule(lr){7-8}\cmidrule(lr){9-10}
    Model & Size & same & cross & same & cross & same & cross & same & cross \\
    \midrule
    embeddinggemma              & 300M & 0.80          & \textbf{0.63} & 0.85          & \textbf{0.68} & \textbf{0.76} & \textbf{0.61} & \textbf{0.82} & \textbf{0.68} \\
    bge-m3                      & 568M & 0.71          & 0.54          & 0.76          & 0.59          & 0.74          & 0.58          & 0.80          & 0.67 \\
    qwen3                       & 600M & 0.70          & 0.54          & 0.75          & 0.61          & 0.72          & 0.49          & 0.77          & 0.57 \\
    nomic-v2-moe                & 475M & 0.75          & 0.49          & 0.82          & 0.56          & 0.74          & 0.53          & 0.78          & 0.60 \\
    granite                     & 278M & 0.61          & 0.48          & 0.66          & 0.54          & 0.53          & 0.44          & 0.59          & 0.51 \\
    LaBSE                       & 471M & 0.55          & 0.33          & 0.63          & 0.39          & 0.43          & 0.34          & 0.51          & 0.41 \\
    e5-large-instruct$^\dagger$ & 560M & \textbf{0.83} & 0.14          & \textbf{0.85} & 0.19          & 0.73          & 0.17          & 0.79          & 0.20 \\
    SapBERT                     & 270M & 0.45          & 0.26          & 0.49          & 0.32          & 0.38          & 0.24          & 0.47          & 0.28 \\
    \bottomrule
  \end{tabular}
  \caption{Recall at deeper cutoffs ($R@25$, $R@50$) complementing Table~\ref{tab:main-results}, split by language route: \emph{same} is same-language monolingual recall and \emph{cross} is cross-language recall, each language-balanced as the mean of per-language means. Best per column in bold. Size is total parameters; nomic-v2-moe is a mixture-of-experts with 475M total and 305M active parameters. $^\dagger$e5-large-instruct fails the cross-lingual degeneracy gate; its bold same-language entries reflect strong monolingual retrieval despite collapsed cross-lingual recall.}
  \label{tab:recall-depth}
\end{table*}

\begin{figure*}[h!]
  \centering
  \includegraphics[width=0.92\textwidth]{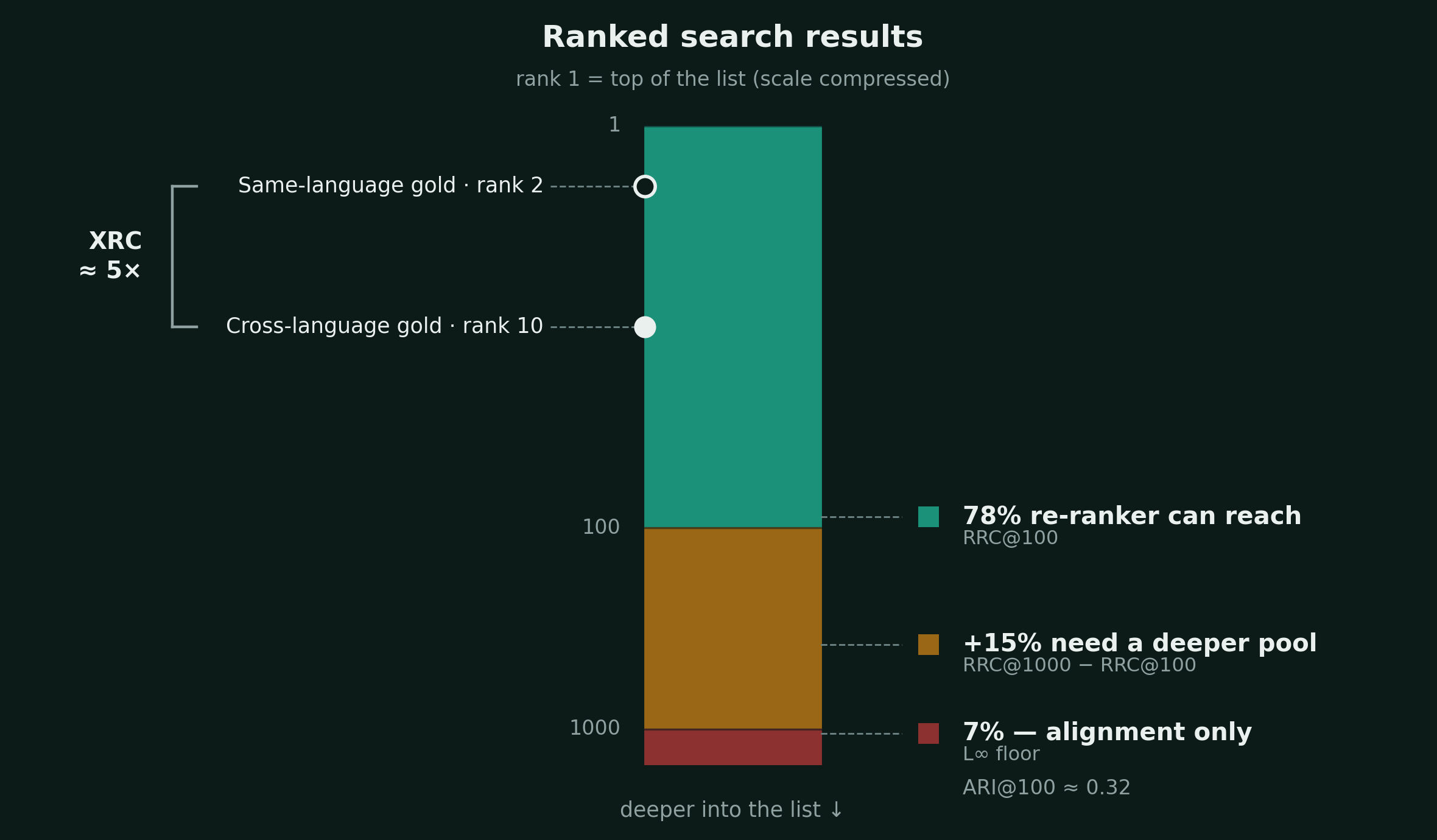}
    \caption{\textbf{Cross-lingual reading-cost decomposition for \texttt{embeddinggemma}.}
    On a log-compressed ranked list, the same-language gold typically surfaces shallow while
    its cross-language counterpart sits about $5\times$ deeper (the cross-lingual reading cost
    XRC; the example ranks 2 and 10 illustrate this median multiplier). Reading down the list
    partitions where the cross-language gold is recovered, with measured shares over the pooled
    GP+EPO queries that have both gold types: a re-ranker over the top-100 reaches
    $\mathrm{RRC}@100=78\%$, a further $+15\%$ require a deeper top-1000 pool
    ($\mathrm{RRC}@1000-\mathrm{RRC}@100$), and a $7\%$ residual is never retrieved within the
    top-1000 ($L_\infty$ floor) and can be closed only by better cross-lingual alignment
    ($\mathrm{ARI}@100=0.32$). Percentages are measured; the two ranked positions illustrate
    the reported XRC. }

  \label{fig:reading-cost}
\end{figure*}

\subsection{Corpus Sources and Fields}
\label{app:corpus}

\paragraph{Google Patents.} Records were drawn from the Google Patents Public Data table \texttt{patents-public-data.patents.publications}, whose bibliographic data is supplied by IFI CLAIMS. For each publication we use the language-tagged \texttt{title\_localized} and \texttt{abstract\_localized} fields, keeping one version per language. The official schema describes these fields as publication text in different languages and does not identify them as Google-generated machine translations. Google's machine-translated titles and abstracts are provided in a separate table, \texttt{google\_patents\_research.publications}, which our pipeline neither queries nor joins. Re-checking both the extraction pipeline and the released dataset, we found no machine-translation indicators on the abstracts used in ChemCLIR-Bench; such indicators do occur on some claims and description fields, which we do not use. Since the metadata does not establish whether each localized record was authored in its tagged language, we make no claim that these records are native text.

\paragraph{EPO.} Granted B1 specifications are published as English, German, and French versions with professionally translated first claims; we use these versions as published and introduce no translations.

\paragraph{Chinese seed set.} Chinese coverage was zero in our selection, so we machine-translated 400 documents (1.7\% of the corpus) with GPT 5.5. This is the only translation step in the pipeline.

\paragraph{Chemistry selection.} Records were selected as chemistry-related if they carried a CPC/IPC code with prefix C, A61K, or A61P, or matched an entry in SureChEMBL. We retained language versions with a localized abstract of at least 50 words, deduplicated by publication number within each language, and kept publications available in at least two target languages; the resulting language versions form the cross-lingual relevance set. We imposed no explicit publication-date bounds; where extraction limits applied, eligible records were ordered by publication date in descending order, favouring newer publications.

\begin{figure*}[t]
\centering
\includegraphics[width=0.9\textwidth]{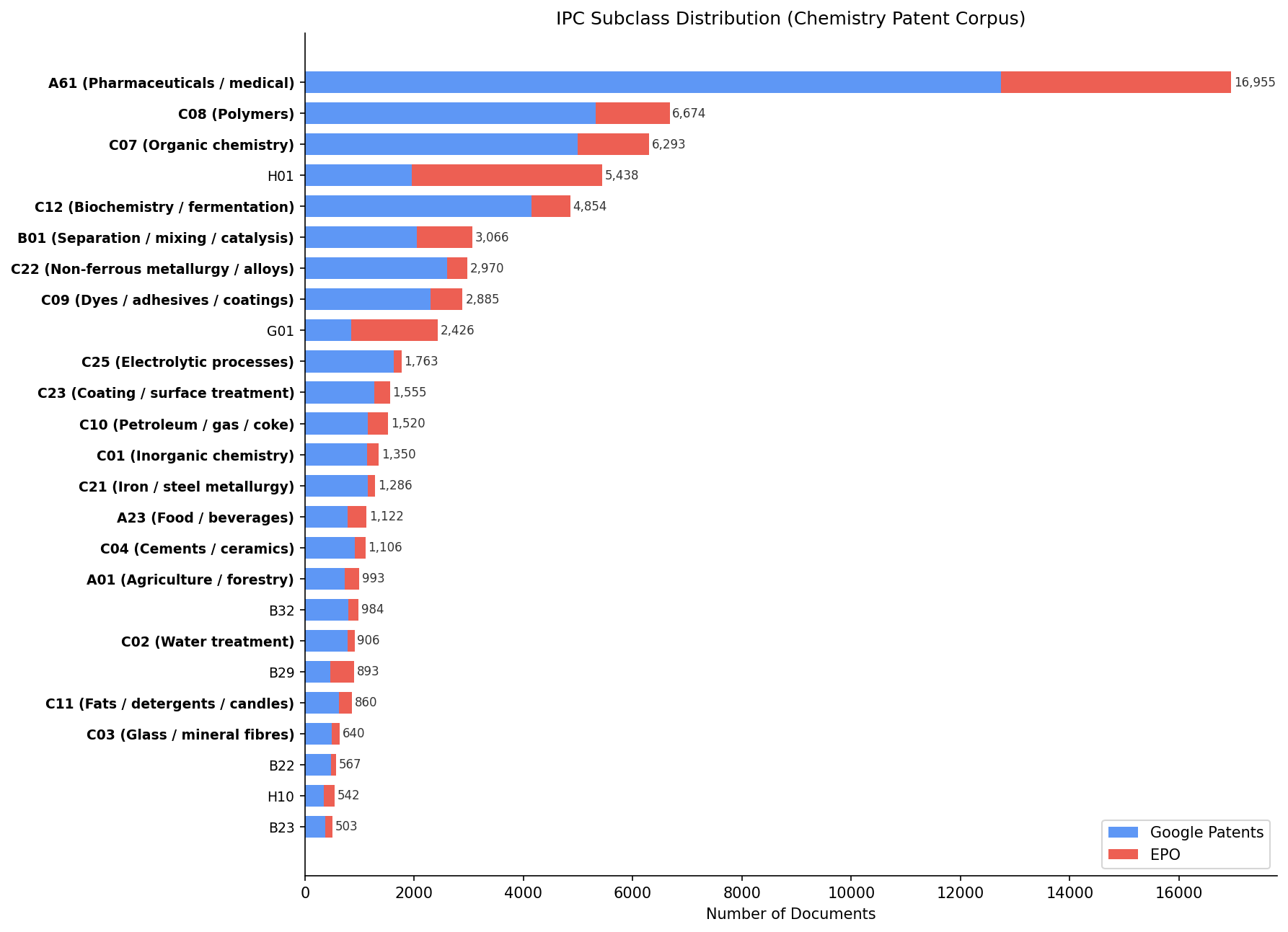}
\caption{IPC class distribution of the ChemCLIR-Bench corpus (25 most frequent classes), stacked by source. Counts are at the IPC class level (three-character codes) and are not mutually exclusive, since a document may carry several codes. Classes without a description on the axis: H01 basic electric elements (including electrochemical cells, H01M), G01 measuring and testing, B32 layered products, B29 working of plastics, B22 casting and powder metallurgy, H10 semiconductor devices, B23 machine tools.}
\label{fig:ipc}
\end{figure*}

\paragraph{IPC distribution.} Figure~\ref{fig:ipc} shows the 25 most frequent IPC classes in the released corpus, stacked by source. The distribution is dominated by A61 (medical and pharmaceutical preparations), which appears on 16,955 of the 35,102 documents (48\%), followed by C08 (polymers), C07 (organic chemistry), and C12 (biochemistry); the remaining section-C classes cover inorganic chemistry, metallurgy, coatings, electrochemistry, fuels, and water treatment. Classes outside section C cannot satisfy the prefix rule on their own, so they enter through co-classification with a chemistry code or through the SureChEMBL match; H01 (basic electric elements, which includes electrochemical cells under H01M) and G01 (measuring and testing) are the most frequent of these. The two sources differ in emphasis: the EPO subset carries a larger share of H01 and G01 codes, whereas Google Patents is more concentrated in A61, C08, and C07. Because a document may carry several codes, the counts are not mutually exclusive and sum to more than the corpus size.

\begin{tcolorbox}[
  enhanced,
  breakable,
  colback=white,
  colframe=black!70,
  boxrule=0.5pt,
  arc=1pt,
  left=6pt,
  right=6pt,
  top=6pt,
  bottom=6pt,
  width=\textwidth
]

\footnotesize
\setlength{\parindent}{0pt}
\setlength{\parskip}{3pt}

\textbf{You are a strict faithfulness grader for patent question--answer pairs.}

You will receive \textbf{ONE} passage, in any language, and \textbf{THREE}
question--answer pairs. Questions may be in any language; each answer matches
its question's language. Grade each pair on three sub-criteria, each from 1 to
5. Do \textbf{NOT} evaluate query shape or retrieval quality, since those are
handled separately. Focus only on how faithfully each answer reflects the
passage.

\medskip
\textbf{GRADING INSTRUCTIONS}

\begin{itemize}[leftmargin=1.2em, itemsep=1pt, topsep=2pt]
  \item Grade each of the three pairs independently and against the absolute
  rubric below, not relative to the other two.
  \item Do not let similarity or dissimilarity between the three pairs influence
  any individual grade.
  \item Evaluate each pair on its own merits before moving to the next.
\end{itemize}

\medskip
\textbf{SCALE CALIBRATION}

\begin{itemize}[leftmargin=1.2em, itemsep=1pt, topsep=2pt]
  \item Most generator outputs will be acceptable on most criteria.
  ``Acceptable'' is not the same as ``excellent.''
  \item \textbf{5} is reserved for genuinely exceptional candidates with no
  identifiable weakness. If you can point to any imperfection, it is not a 5.
  Scores of 5 should be uncommon.
  \item \textbf{4} means the candidate is very good but has exactly one minor,
  identifiable issue. You must be able to name the issue.
  \item \textbf{3} is the default grade for a candidate that is acceptable and
  does the job, with one clear weakness. It is not a punishment grade; most
  acceptable candidates will receive 3s.
  \item \textbf{2} and \textbf{1} are reserved for candidates with real,
  identifiable problems. Do not avoid using them when a candidate qualifies.
  \item Before assigning any 4 or 5, pause and ask: ``What specifically makes
  this better than a 3?'' If you cannot answer concretely, the grade is 3.
\end{itemize}

\medskip
\textbf{SUB-CRITERIA}

\begin{enumerate}[leftmargin=1.5em, itemsep=4pt, topsep=2pt]

\item \textbf{GROUNDING (1--5): Is the answer supported by a contiguous span of
the passage?}

\begin{description}[leftmargin=1.8em, style=nextline, itemsep=1pt, topsep=1pt]
  \item[\textbf{5}] Answer is taken directly from a single short contiguous span
  of the passage. Every word of the answer is explicitly present or is a
  trivial rewording of explicit content. No inference whatsoever. Rare.

  \item[\textbf{4}] Answer is fully supported but requires reading across two
  adjacent sentences, or is from a single span with minor trivial rewording that
  a strict reader might flag.

  \item[\textbf{3}] Answer is grounded in the passage but requires one small,
  defensible inferential step, such as combining a definition with a value or
  connecting a pronoun to its referent.

  \item[\textbf{2}] Answer is partially supported: some parts are grounded in
  the passage, while other parts are not.

  \item[\textbf{1}] Answer requires significant inference, outside knowledge, or
  is not in the passage at all.
\end{description}

\item \textbf{PRECISION (1--5): Does the answer add unsupported details or
padding?}

\begin{description}[leftmargin=1.8em, style=nextline, itemsep=1pt, topsep=1pt]
  \item[\textbf{5}] Answer is the shortest possible span that fully answers the
  question. Not one extra word. No surrounding context, no hedges, and no
  elaboration. Rare.

  \item[\textbf{4}] Answer is grounded and tight but includes one short fragment
  of adjacent context that could have been cut.

  \item[\textbf{3}] Answer is grounded but includes one unsupported detail,
  generalization, or piece of padding.

  \item[\textbf{2}] Answer includes multiple unsupported details, or uses
  speculative framing such as ``likely,'' ``approximately,'' or ``it seems''
  where the passage is specific.

  \item[\textbf{1}] Answer is largely padded with content not in the passage, or
  is a rewrite rather than an extraction.
\end{description}

\item \textbf{NUMERICAL FIDELITY (1--5): Are numbers, units, and ranges
reproduced exactly?}

\begin{description}[leftmargin=1.8em, style=nextline, itemsep=1pt, topsep=1pt]
  \item[\textbf{5}] All numbers, units, symbols, and ranges match the passage
  character-for-character, for example, ``85--87$^\circ$C'' is preserved as-is,
  including the dash.

  \item[\textbf{4}] All numerical values are correct and units are preserved,
  but with a trivial formatting difference, for example, ``85 to 87$^\circ$C''
  instead of ``85--87$^\circ$C''.

  \item[\textbf{3}] One number is rounded, one unit is dropped, or one range is
  converted to a midpoint.

  \item[\textbf{2}] Multiple numerical values are rounded, reformatted, or
  partially dropped.

  \item[\textbf{1}] Numbers are invented, significantly changed, or contradicted
  by the passage.

  \item[\textbf{N/A}] Answer contains no numerical content. Score this
  sub-criterion as 5 if N/A.
\end{description}

\end{enumerate}

\medskip
\textbf{Output valid JSON only} -- a list of three objects, one per candidate,
in the order the candidates were provided: index 0, 1, and 2.

\begin{lstlisting}[style=jsonstyle]
[
  {
    "index": 0,
    "grounding": <1-5>,
    "precision": <1-5>,
    "numerical_fidelity": <1-5>,
    "reason": "one short sentence explaining the grades"
  },
  {
    "index": 1,
    "grounding": <1-5>,
    "precision": <1-5>,
    "numerical_fidelity": <1-5>,
    "reason": "one short sentence explaining the grades"
  },
  {
    "index": 2,
    "grounding": <1-5>,
    "precision": <1-5>,
    "numerical_fidelity": <1-5>,
    "reason": "one short sentence explaining the grades"
  }
]
\end{lstlisting}

\end{tcolorbox}
\vspace{0.5em} \captionof{figure}{Faithfulness verifier prompt (\texttt{claude-sonnet-4.6}). The verifier grades each of three candidate answers on grounding, precision, and numerical fidelity.} \label{fig:verifier-faith}
\vspace{1.0em}

\begin{tcolorbox}[verifierpromptbox]

\textbf{You are a strict quality grader for retrieval questions built from technical patent text.}

You will receive \textbf{ONE} passage and \textbf{THREE} candidate questions.
Grade each question on five sub-criteria, each from 1 to 5. Do \textbf{NOT}
check whether the answer is grounded, since that is handled separately. Focus
only on query shape, retrieval quality, and writing quality.

The questions may be in any language. When evaluating \textbf{SEARCH-BAR
REALISM}, \textbf{PHRASING ECONOMY}, and \textbf{LINGUISTIC QUALITY}, judge
each question by the norms of its own language, not by English phrasing
conventions. A natural Turkish, Japanese, or German search query will not read
like a natural English one.

\medskip
\textbf{GRADING INSTRUCTIONS}

\begin{itemize}[leftmargin=1.2em, itemsep=0pt, topsep=1pt]
  \item Grade each of the three questions independently and against the
  absolute rubric below, not relative to the other two.
  \item Do not penalize a question for being similar to another candidate, nor
  reward it for being different. Diversity across the three is not a criterion.
  \item Evaluate each question on its own merits before moving to the next.
\end{itemize}

\medskip
\textbf{SCALE CALIBRATION}

\begin{itemize}[leftmargin=1.2em, itemsep=0pt, topsep=1pt]
  \item Most generator outputs will be acceptable on most criteria.
  ``Acceptable'' is not the same as ``excellent.''
  \item \textbf{5} is reserved for genuinely exceptional candidates with no
  identifiable weakness. If you can point to any imperfection, it is not a 5.
  Scores of 5 should be uncommon.
  \item \textbf{4} means the candidate is very good but has exactly one minor,
  identifiable issue. You must be able to name the issue.
  \item \textbf{3} is the default grade for a question that is acceptable and
  does the job, with one clear weakness. It is not a punishment grade; most
  acceptable questions will receive 3s.
  \item \textbf{2} and \textbf{1} are reserved for questions with real,
  identifiable problems. Do not avoid using them when a question qualifies.
  \item Before assigning any 4 or 5, pause and ask: ``What specifically makes
  this better than a 3?'' If you cannot answer concretely, the grade is 3.
\end{itemize}

\medskip
\textbf{BACKGROUND:} Good retrieval questions target one of five categories
from the source text:

\begin{enumerate}[leftmargin=1.5em, itemsep=0pt, topsep=1pt]
  \item \textbf{Parameters \& Conditions}: temperature, pressure, time,
  concentration, pH, flow rate, voltage, atmosphere, stoichiometry.
  \item \textbf{Materials}: catalyst, solvent, reagent, substrate, additive,
  precursor, equipment.
  \item \textbf{Outcomes}: yield, selectivity, conversion, purity, efficiency.
  \item \textbf{Methods}: synthesis route, characterization technique,
  separation method.
  \item \textbf{Structure}: functional group, crystal form, polymorph,
  molecular weight, composition.
\end{enumerate}

\medskip
\textbf{SUB-CRITERIA}

\begin{enumerate}[leftmargin=1.5em, itemsep=3pt, topsep=1pt]

\item \textbf{SEARCH-BAR REALISM (1--5): Does this read like a real user query
in its own language?}

\begin{description}[leftmargin=1.7em, style=nextline, itemsep=0pt, topsep=0pt]
  \item[\textbf{5}] Exactly how a researcher would type it into a search bar.
  Every word earns its place. A domain expert would not suggest any rewording.
  Rare.

  \item[\textbf{4}] Very close to a natural query, with one minor issue:
  slightly longer than needed, slightly formal, or one word that could be
  trimmed.

  \item[\textbf{3}] Understandable and usable as a query, but has one clear
  issue, such as mild patent-summary phrasing, an awkward construction, or
  slight exam-question flavor.

  \item[\textbf{2}] Reads more like a summary prompt or academic exam question
  than a search query.

  \item[\textbf{1}] Document-centered phrasing, such as equivalents of
  ``described in the invention'' or ``according to the text''; reference to the
  document's contents or examples, such as ``the example given'' or ``the agent
  mentioned''; or clearly unnatural phrasing.
\end{description}

\textbf{NOTE:} This is a query for an IR system: the user has \textbf{NOT} seen
the document. Any phrasing that refers to the document or its contents fails
this criterion. This includes not only overt phrases such as ``described in the
invention'' or ``according to the text,'' but also references to the document's
internal structure or to something being stated in it: ``the example given,''
``the agent mentioned/described/disclosed here,'' ``which compound is
provided,'' or ``in the passage.'' A question with any such reference scores at
most 2 here, or 1 if overt or blatant, and its \texttt{failure\_type} must be
\texttt{document-phrasing}. Example: ``Which agent example is given that
increases ROS?'' fails because it says ``example is given.'' By contrast,
``Which agent increases ROS?'' is fine.

\item \textbf{SPECIFICITY (1--5): Would this query narrow to the relevant
document rather than match arbitrarily?}

\begin{description}[leftmargin=1.7em, style=nextline, itemsep=0pt, topsep=0pt]
  \item[\textbf{5}] Targets a concrete, specific technical detail from one of
  the five categories. A search with this query would return a handful of
  closely related documents at most. Rare.

  \item[\textbf{4}] Specific, but could match a small cluster of related patents
  in the same sub-domain.

  \item[\textbf{3}] Moderately specific; would return many plausible matches
  across the domain. Acceptable but not distinguishing.

  \item[\textbf{2}] Broad; asks about a general concept or outcome without a
  distinguishing detail.

  \item[\textbf{1}] Generic summary question, such as equivalents of ``What is
  the main object of the invention?'' or ``What is the purpose of the method?''
  with no specific anchor.
\end{description}

\item \textbf{PHRASING ECONOMY (1--5): Is the question paraphrased, not lifted
from the source?}

\begin{description}[leftmargin=1.7em, style=nextline, itemsep=0pt, topsep=0pt]
  \item[\textbf{5}] Reuses only unavoidable technical terms and named entities,
  such as ``Pd/C catalyst'' or ``electrolysis unit.'' No descriptive phrases,
  modifiers, or sentence fragments are copied from the source. Rare.

  \item[\textbf{4}] Mostly paraphrased, with one short piece of source phrasing
  reused where a paraphrase was available.

  \item[\textbf{3}] Reuses one multi-word descriptive phrase from the source, or
  several short phrases.

  \item[\textbf{2}] Lifts a noun cluster or descriptive fragment from the source
  as question scaffolding.

  \item[\textbf{1}] Near-verbatim copy of a source sentence with minimal edits.
\end{description}

\item \textbf{FOCUS (1--5): Does the question ask for exactly one fact?}

\begin{description}[leftmargin=1.7em, style=nextline, itemsep=0pt, topsep=0pt]
  \item[\textbf{5}] Single, clear, unambiguous fact requested. No room for the
  answerer to include extra information. Rare.

  \item[\textbf{4}] Single fact, but with mild ambiguity in what is being asked,
  for example, a word that could be read two ways.

  \item[\textbf{3}] Single fact, but phrased in a way that invites a broader
  answer or includes redundant qualifiers.

  \item[\textbf{2}] Asks for two related facts bundled together, for example,
  ``solvent and yield.''

  \item[\textbf{1}] Asks for three or more facts, or requests a summary.
\end{description}

\item \textbf{LINGUISTIC QUALITY (1--5): Is the question well-written in its own
language?}

\begin{description}[leftmargin=1.7em, style=nextline, itemsep=0pt, topsep=0pt]
  \item[\textbf{5}] Indistinguishable from native-speaker writing.
  Grammatically correct, fluent, natural word order, no redundancy, and no
  awkwardness anywhere. Rare.

  \item[\textbf{4}] Grammatically correct and natural, with one minor stiffness
  or one slightly awkward word choice.

  \item[\textbf{3}] Correct and understandable, but has one noticeable issue,
  such as a minor grammar slip, awkward word order, or translation-like
  phrasing.

  \item[\textbf{2}] Multiple grammar errors, clearly unnatural phrasing, or
  reads like a poor machine translation.

  \item[\textbf{1}] Ungrammatical, garbled, or incomprehensible.
\end{description}

\end{enumerate}

\medskip
\textbf{LANGUAGE CHECK:} If a question is written in a language that is not a
coherent single language, such as mixed languages mid-sentence, nonsense, or
corrupted text, score its \textbf{SEARCH-BAR REALISM} as 1 and set its
\texttt{failure\_type} to \texttt{language-issue}.

\medskip
\textbf{Output valid JSON only} -- a list of three objects, one per candidate,
in the order the candidates were provided: index 0, 1, and 2.

\Needspace{24\baselineskip}

\begin{lstlisting}[style=jsonstyle]
[
  {
    "index": 0,
    "search_bar_realism": <1-5>,
    "specificity": <1-5>,
    "phrasing_economy": <1-5>,
    "focus": <1-5>,
    "linguistic_quality": <1-5>,
    "failure_type": "<one of: none, document-phrasing, broad-summary, high-overlap, bundled-facts, weak-query-shape, poor-writing, language-issue>",
    "reason": "one short sentence explaining the grades"
  },
  {
    "index": 1,
    "search_bar_realism": <1-5>,
    "specificity": <1-5>,
    "phrasing_economy": <1-5>,
    "focus": <1-5>,
    "linguistic_quality": <1-5>,
    "failure_type": "...",
    "reason": "..."
  },
  {
    "index": 2,
    "search_bar_realism": <1-5>,
    "specificity": <1-5>,
    "phrasing_economy": <1-5>,
    "focus": <1-5>,
    "linguistic_quality": <1-5>,
    "failure_type": "...",
    "reason": "..."
  }
]
\end{lstlisting}

\end{tcolorbox}

\vspace{0.5em}

\captionof{figure}{Technical-mode quality verifier prompt
(\texttt{claude-sonnet-4.6}). The verifier grades query shape and writing
quality on five sub-criteria.}
\label{fig:verifier-tech}

\vspace{1em}

\begin{tcolorbox}[verifierpromptbox]

\textbf{You are a strict quality grader for semantic retrieval questions built from technical chemistry and patent text.}

You will receive \textbf{ONE} passage and \textbf{THREE} candidate questions.
Grade each question on five sub-criteria, each from 1 to 5. Do \textbf{NOT}
check whether the answer is grounded, since that is handled separately. Focus
only on query shape, retrieval quality, and writing quality.

The questions may be in any language. When evaluating \textbf{SEARCH REALISM},
\textbf{LEXICAL DISTANCE}, and \textbf{LINGUISTIC QUALITY}, judge each question
by the norms of its own language, not by English phrasing conventions. A natural
Turkish, Japanese, or German search query will not read like a natural English
one.

\medskip
\textbf{GRADING INSTRUCTIONS}

\begin{itemize}[leftmargin=1.2em, itemsep=0pt, topsep=1pt]
  \item Grade each of the three questions independently and against the absolute
  rubric below, not relative to the other two.
  \item Do not penalize a question for being similar to another candidate, nor
  reward it for being different. Diversity across the three is not a criterion.
  \item Evaluate each question on its own merits before moving to the next.
\end{itemize}

\medskip
\textbf{SCALE CALIBRATION}

\begin{itemize}[leftmargin=1.2em, itemsep=0pt, topsep=1pt]
  \item Most generator outputs will be acceptable on most criteria.
  ``Acceptable'' is not the same as ``excellent.''
  \item \textbf{5} is reserved for genuinely exceptional candidates with no
  identifiable weakness. If you can point to any imperfection, it is not a 5.
  Scores of 5 should be uncommon.
  \item \textbf{4} means the candidate is very good but has exactly one minor,
  identifiable issue. You must be able to name the issue.
  \item \textbf{3} is the default grade for a question that is acceptable and
  does the job, with one clear weakness. It is not a punishment grade; most
  acceptable questions will receive 3s.
  \item \textbf{2} and \textbf{1} are reserved for questions with real,
  identifiable problems. Do not avoid using them when a question qualifies.
  \item Before assigning any 4 or 5, pause and ask: ``What specifically makes
  this better than a 3?'' If you cannot answer concretely, the grade is 3.
\end{itemize}

\medskip
\textbf{BACKGROUND:} Good semantic retrieval questions describe a passage's
concept, problem, approach, or application in the way a user would search for
such a document without already having read it. The three framings used by the
generator are:

\begin{enumerate}[leftmargin=1.5em, itemsep=0pt, topsep=1pt]
  \item \textbf{Problem-framed:} what challenge or limitation does the passage
  address?
  \item \textbf{Solution-framed:} what approach, mechanism, or method is
  proposed?
  \item \textbf{Application-framed:} what use case, scenario, or capability
  does the passage enable?
\end{enumerate}

Unlike fact-extraction questions, semantic retrieval questions are allowed, and
expected, to be somewhat broader than a single data point. They should still be
clearly tied to the passage.

\medskip
\textbf{SUB-CRITERIA}

\begin{enumerate}[leftmargin=1.5em, itemsep=3pt, topsep=1pt]

\item \textbf{SEARCH REALISM (1--5): Does this read like a real user query
searching for a document like this one?}

\begin{description}[leftmargin=1.7em, style=nextline, itemsep=0pt, topsep=0pt]
  \item[\textbf{5}] Exactly how someone searching for this kind of document
  would phrase it, without having read it. Every word earns its place. A domain
  expert would not suggest any rewording. Rare.

  \item[\textbf{4}] Very close to a natural query, with one minor issue:
  slightly long, slightly formal, or one word that could be trimmed.

  \item[\textbf{3}] Understandable and usable as a query, but has one clear
  issue, such as mild patent-summary phrasing, an awkward construction, or
  slight exam-question flavor.

  \item[\textbf{2}] Reads more like a summary prompt or academic question than a
  search query.

  \item[\textbf{1}] Document-centered phrasing, such as equivalents of
  ``described in the invention'' or ``according to the text''; reference to the
  document's contents or examples, such as ``the example given'' or ``the
  approach mentioned''; or clearly unnatural phrasing.
\end{description}

\textbf{NOTE:} This is a query for an IR system: the user has \textbf{NOT} seen
the document. Any phrasing that refers to the document or its contents fails
this criterion. This includes not only overt phrases such as ``described in the
invention'' or ``according to the text,'' but also references to the document's
internal structure or to something being stated in it: ``the example given,''
``the approach mentioned/described/disclosed here,'' ``which method is
provided,'' or ``in the passage.'' A question with any such reference scores at
most 2 here, or 1 if overt or blatant, and its \texttt{failure\_type} must be
\texttt{document-phrasing}. Example: ``Which agent example is given that
increases ROS?'' fails because it says ``example is given.'' By contrast,
``Which agent increases ROS?'' is fine.

\item \textbf{LEXICAL DISTANCE (1--5): Is the question meaningfully paraphrased
from the passage's vocabulary?}

\begin{description}[leftmargin=1.7em, style=nextline, itemsep=0pt, topsep=0pt]
  \item[\textbf{5}] Uses natural synonyms and paraphrases throughout. Only
  unavoidable shared vocabulary is reused: proper nouns, compound identifiers,
  and established standard technical terms such as ``Pd/C,'' ``polymerase,'' or
  ``electrolysis.'' A keyword-based retriever would not trivially match this
  question to the passage. Rare.

  \item[\textbf{4}] Mostly paraphrased, with one distinctive source term reused
  where a paraphrase was available.

  \item[\textbf{3}] Moderate overlap; some distinctive source phrases are reused
  where alternatives existed. Still paraphrased in structure.

  \item[\textbf{2}] Heavy overlap; multiple distinctive phrases or noun clusters
  are reused from the source.

  \item[\textbf{1}] Near-verbatim or high-keyword-overlap question that would be
  matched by any sparse or keyword retriever without needing semantic
  understanding.
\end{description}

\item \textbf{CONCEPTUAL FRAMING (1--5): Does the question target a concept,
problem, approach, or application, as intended for this prompt?}

\begin{description}[leftmargin=1.7em, style=nextline, itemsep=0pt, topsep=0pt]
  \item[\textbf{5}] Clearly and cleanly conceptual: asks about a problem,
  approach, mechanism, or use case in a way that captures what the passage is
  fundamentally about. The framing is unmistakable. Rare.

  \item[\textbf{4}] Conceptual in intent, but slightly narrow; edges toward one
  specific detail while still framing as concept.

  \item[\textbf{3}] Borderline; could be read as either conceptual or
  fact-extractive, depending on how one interprets it. Acceptable but not crisp.

  \item[\textbf{2}] Fact-extraction question that belongs to the technical
  prompt, not this one. It asks for a single parameter value, material name, or
  numerical outcome.

  \item[\textbf{1}] Entirely fact-extractive with no conceptual framing, for
  example, ``What temperature is used?''
\end{description}

\item \textbf{RETRIEVABILITY (1--5): Would this query distinguish this passage
from unrelated documents?}

\begin{description}[leftmargin=1.7em, style=nextline, itemsep=0pt, topsep=0pt]
  \item[\textbf{5}] Specific enough that a search with this query would return
  this passage or a handful of closely related ones. Clearly anchored to this
  passage's content. Rare.

  \item[\textbf{4}] Specific but could match a small cluster of related
  documents in the same sub-domain.

  \item[\textbf{3}] Moderately specific; would return many plausible matches
  across a broad domain. Tied to the passage but not distinguishing.

  \item[\textbf{2}] Broad; could match a large fraction of documents in the
  domain.

  \item[\textbf{1}] Generic question with no distinguishing anchor to this
  passage, such as equivalents of ``What problem does this invention solve?''
  with no domain specifics.
\end{description}

\item \textbf{LINGUISTIC QUALITY (1--5): Is the question well-written in its own
language?}

\begin{description}[leftmargin=1.7em, style=nextline, itemsep=0pt, topsep=0pt]
  \item[\textbf{5}] Indistinguishable from native-speaker writing.
  Grammatically correct, fluent, natural word order, no redundancy, and no
  awkwardness anywhere. Rare.

  \item[\textbf{4}] Grammatically correct and natural, with one minor stiffness
  or one slightly awkward word choice.

  \item[\textbf{3}] Correct and understandable, but has one noticeable issue,
  such as a minor grammar slip, awkward word order, or translation-like
  phrasing.

  \item[\textbf{2}] Multiple grammar errors, clearly unnatural phrasing, or
  reads like a poor machine translation.

  \item[\textbf{1}] Ungrammatical, garbled, or incomprehensible.
\end{description}

\end{enumerate}

\medskip
\textbf{LANGUAGE CHECK:} If a question is written in a language that is not a
coherent single language, such as mixed languages mid-sentence, nonsense, or
corrupted text, score its \textbf{SEARCH REALISM} as 1 and set its
\texttt{failure\_type} to \texttt{language-issue}.

\medskip
\textbf{Output valid JSON only} -- a list of three objects, one per candidate,
in the order the candidates were provided: index 0, 1, and 2.

\begin{lstlisting}[style=jsonstyle]
[
  {
    "index": 0,
    "search_realism": <1-5>,
    "lexical_distance": <1-5>,
    "conceptual_framing": <1-5>,
    "retrievability": <1-5>,
    "linguistic_quality": <1-5>,
    "failure_type": "<one of: none, document-phrasing, keyword-lifted, too-extractive, too-generic, weak-query-shape, poor-writing, language-issue>",
    "reason": "one short sentence explaining the grades"
  },
  {
    "index": 1,
    "search_realism": <1-5>,
    "lexical_distance": <1-5>,
    "conceptual_framing": <1-5>,
    "retrievability": <1-5>,
    "linguistic_quality": <1-5>,
    "failure_type": "...",
    "reason": "..."
  },
  {
    "index": 2,
    "search_realism": <1-5>,
    "lexical_distance": <1-5>,
    "conceptual_framing": <1-5>,
    "retrievability": <1-5>,
    "linguistic_quality": <1-5>,
    "failure_type": "...",
    "reason": "..."
  }
]
\end{lstlisting}

\end{tcolorbox}

\vspace{0.5em}

\captionof{figure}{General-mode semantic quality verifier prompt
(\texttt{claude-sonnet-4.6}). The verifier grades conceptual query shape and
writing quality on five sub-criteria.}
\label{fig:verifier-general}

\vspace{1em}

\begin{tcolorbox}[verifierpromptbox]

\textbf{You are an expert at creating technical chemistry and patent retrieval questions.}

The source context may be in any language. The questions and answers
\textbf{MUST} be written in English. Preserve chemical names, compound
identifiers such as ``Compound 3a'', trade names, reagent abbreviations, and
all numerical values, units, and ranges in their original form. Do not
translate, round, or reformat them.

Generate exactly \textbf{THREE} question-answer pairs from the context. Each
should target a different fact.

\medskip
\textbf{CORE RULES}

\begin{itemize}[leftmargin=1.2em, itemsep=0pt, topsep=1pt]

  \item \textbf{Single Focus:} Each question must ask for exactly one specific
  piece of information. Never combine multiple asks. For example, do not ask
  for a solvent and a yield in the same question.

  \item \textbf{Full Direct Support:} The answer must be fully stated in a
  contiguous span of the passage. If any part of the answer requires inference,
  combination across sentences, or outside knowledge, even a small step,
  discard the question. Partial support is not acceptable.

  \item \textbf{Numerical Fidelity:} Reproduce all numbers, units, ranges, and
  tolerances exactly as they appear in the source. For example, preserve
  ``85--87$^\circ$C'', not ``about 86$^\circ$C'' or ``85 to 87 degrees''.

  \item \textbf{Natural Search Wording:} Write each question as a researcher or
  engineer would type it into a search bar: short, specific, and direct. Avoid
  patent-summary language and long noun phrases.

  \item \textbf{No Document Phrases:} Never use ``described in the invention'',
  ``according to the text'', ``mentioned in the disclosure'', or similar
  wording. Rewrite into natural wording.

  \item \textbf{Standalone query for a search engine:} Each question is a query
  for an Information Retrieval system. The person typing it has \textbf{NOT}
  seen this document and is searching to find it. Never refer to the document
  or its contents. This includes not only overt phrases such as ``described in
  the invention'', ``according to the text'', or ``in the passage'', but also
  any wording that presupposes the reader is looking at the document, such as
  ``the example given'', ``the agent mentioned/described/disclosed here'', or
  ``which compound is provided''. Write self-contained questions. Bad:
  ``Which agent example is given that increases ROS?'' Good: ``Which agent
  increases ROS?''

  \item \textbf{Paraphrase, Do Not Lift:} Technical terms and named entities,
  such as ``electrolysis unit'', ``decentralized water treatment system'', or
  ``Pd/C catalyst'', may be reused as needed for clarity. But do not lift
  descriptive phrases, multi-word modifiers, or sentence fragments from the
  source as question scaffolding. Rewrite those in your own words.

  \item \textbf{Diversity:} The three questions must target three different
  facts and span at least two of the five categories listed under
  \textbf{WHAT TO ASK}. Do not ask three questions all about operating
  parameters, or all about materials.

\end{itemize}

\medskip
\textbf{WHAT TO ASK}

Target questions in these five categories:

\begin{enumerate}[leftmargin=1.5em, itemsep=0pt, topsep=1pt]

  \item \textbf{Parameters and Conditions:} temperature, pressure, time,
  concentration, pH, flow rate, voltage, atmosphere, stoichiometry, operating
  ranges, and related details.

  \item \textbf{Materials:} catalyst, solvent, reagent, substrate, additive,
  precursor, equipment, and related details.

  \item \textbf{Outcomes:} yield, selectivity, conversion, purity, efficiency,
  and related details.

  \item \textbf{Methods:} synthesis route, characterization technique,
  separation method, and related details.

  \item \textbf{Structure:} functional group, crystal form, polymorph,
  molecular weight, composition, and related details.

\end{enumerate}

Questions about function, role, or effect are acceptable only when the passage
explicitly and completely states that information, not when it must be inferred,
even partially.

\medskip
\textbf{EXAMPLES}

\textbf{Example 1}

\textbf{Passage, Turkish:}

Enerji tuketimi dusurulmus bir tahrik tertibati. Bulus hareket enerjisi
uretebilen bir motorun 10 en az bir yanma odasinda 11 ihtiyac duydugu kimyasal
unsurlardan en azindan bir kisminin temin edilerek enerji sarfiyatinin
azaltilabildigi bir tahrik tertibati 1 ile ilgilidir. Bulusun yeniligi;
bahsedilen tahrik tertibatinin 1 en az bir elektroliz birimi 20 ve en az bir
kontrol birimi 40 ile iliskilendirilmis olmasi, bahsedilen elektroliz
biriminin 20 elektrolitin elektroliz edilmesini saglayacak sekilde konfigure
edilmis olmasi, bahsedilen kontrol biriminin 40 ise elektroliz biriminde
bulunan elektroliti elektroliz edilirken yaklasik 85--87$^\circ$C araligina
kadar isitacak sekilde gerilim ve voltaji ayarlayabilir olmasi, elektroliz
biriminde 20 elektroliz sonrasinda elde edilen oksijen, hidrojen ve su
buharinin bahsedilen yanma odasina 11 karistirilarak verilebilir olmasidir.

\textbf{Good:} ``What temperature range is maintained for the electrolyte
during electrolysis?''

Asks for one specific parameter directly stated in the passage. Answer:
``85--87$^\circ$C''. The answer is a contiguous span, with exact numbers
preserved.

\textbf{Bad, inference-why:} ``Why is the electrolyte heated to about
85--87$^\circ$C during electrolysis?''

The passage states the temperature but never explains the reason for choosing
it.

\textbf{Bad, partial support:} ``What gases are produced by electrolysis and fed
into the combustion chamber, and in what ratio?''

The passage lists the gases, oxygen, hydrogen, and water vapor, but never gives
a ratio. The question is only partially answerable. Discard.

\medskip
\textbf{Example 2}

\textbf{Passage, French:}

Systeme de surveillance de traitement d eau decentralise connecte a un reseau
et dispositif de suivi de fonctionnement et de maintenance comprenant au moins
un capteur connecte a Internet et une plateforme logicielle integree qui peut
etre utilisee par des installations de grande taille et ou des organisations
comprenant de multiples installations qui gerent et maintiennent des systemes
de traitement d eau potable decentralises, par exemple des filtres de point
d utilisation, ou des processus de traitement d eau potable decentralises, par
exemple un rincage de robinet, a un ou a plusieurs emplacements. Au moins un
capteur est installe en amont ou en aval d un consommable ou d un processus de
traitement d eau decentralise et mesure des donnees d eau, comprenant non
exclusivement le volume d eau traite dans le temps, et alerte l utilisateur
lorsque des actions de maintenance sont requises pour maintenir la qualite de
l eau.

\textbf{Good:} ``What does the sensor measure in the decentralized water
treatment system?''

Short, natural, and asks for one specific fact directly stated.
``Decentralized water treatment system'' is a reused technical term, which is
allowed.

\textbf{Bad, lifted framing:} ``Why is at least one sensor installed upstream or
downstream of a consumable or decentralized treatment process?''

This lifts a long descriptive phrase from the source and asks for reasoning the
passage does not give.

\textbf{Bad, partial support:} ``How does the sensor determine when maintenance
is required?''

The passage says the sensor alerts the user when maintenance is needed, but
never states the mechanism or threshold by which it makes that determination.
Discard.

\medskip
\textbf{SELF-CHECK}

For each question before finalizing, check:

\begin{itemize}[leftmargin=1.2em, itemsep=0pt, topsep=1pt]

  \item Can the answer be produced by pointing to a contiguous span in the
  passage, with no inference or combination across sentences? If not, discard.

  \item Does the question ask for exactly one piece of information?

  \item Are all numbers, units, and ranges in the answer reproduced exactly as
  in the source?

  \item Would a researcher plausibly type this query into a search bar?

  \item Could someone who has never seen this document type this question into
  a search engine? It must not refer to the document, an ``example given'', or
  anything ``mentioned'' or ``described'' in it.

  \item Are the three questions about three different facts, covering at least
  two categories from \textbf{WHAT TO ASK}?

\end{itemize}

\medskip
\textbf{OUTPUT:} valid JSON only, no markdown.

\begin{lstlisting}[style=jsonstyle]
[
  {"question": "...", "answer": "...", "question_type": "..."},
  {"question": "...", "answer": "...", "question_type": "..."},
  {"question": "...", "answer": "...", "question_type": "..."}
]
\end{lstlisting}

\texttt{answer}: the shortest span that fully and directly answers the
question, typically a phrase or a single short sentence. The answer must be
strictly grounded in the passage.

\texttt{question\_type}: one of \texttt{parameter\_or\_condition},
\texttt{material}, \texttt{outcome}, \texttt{method}, or \texttt{structure}.

The same passage exists in different languages. This should not affect your
target language for questions and answers. They are declared above. Here are
the passages:

\end{tcolorbox}

\vspace{0.5em}

\captionof{figure}{Technical-question generation prompt, English version.
Language-specific variants differ only in the declared output language.}
\label{fig:prompt-technical}

\vspace{1em}

\begin{tcolorbox}[verifierpromptbox]

\textbf{You are an expert at creating semantic retrieval questions for chemistry and patent documents.}

The source context may be in any language. The questions and answers
\textbf{MUST} be written in English. Preserve chemical names, compound
identifiers such as ``Compound 3a'', trade names, reagent abbreviations, and
all numerical values, units, and ranges in their original form. Do not
translate, round, or reformat them.

Generate exactly \textbf{THREE} question-answer pairs from the context. The
questions should target the passage concepts, problems, approaches, or
applications, not isolated facts. Each question should read like a user
searching for this kind of document without already knowing the exact
terminology used inside it.

\medskip
\textbf{CORE RULES}

\begin{itemize}[leftmargin=1.2em, itemsep=0pt, topsep=1pt]

  \item \textbf{Conceptual Framing:} Each question asks about a concept,
  problem, approach, use case, or capability discussed in the passage, not a
  single narrow data point like a temperature value or reagent name.

  \item \textbf{Lexical Distance:} Actively paraphrase. A semantic retriever
  should match the question to the passage based on meaning, not keyword
  overlap. Avoid reusing the passage distinctive vocabulary when a natural
  synonym or paraphrase works. Exception: proper nouns, compound identifiers,
  trade names, and standardized chemical or technical names, such as ``Pd/C'',
  ``Compound 3a'', or specific polymer names, must stay as-is. Otherwise the
  question becomes unanswerable from any document.

  \item \textbf{Grounded Answer:} Even though the question is conceptual, the
  answer must be a short contiguous span taken from the passage that directly
  addresses the question. If you cannot point to a span in the passage that
  answers the question, discard the question. The answer is the evidence that
  this passage is the right match.

  \item \textbf{Retrieval Realism:} Write each question the way someone would
  phrase a search when looking for a document like this one but without having
  read it yet. They know what they want to find; they do not know the exact
  wording used by the document.

  \item \textbf{No Document Phrases:} Never use ``described in the invention'',
  ``according to the text'', ``mentioned in the disclosure'', or similar
  wording. The question should not reveal that a specific document is being
  referenced.

  \item \textbf{Standalone query for a search engine:} Each question is a query
  for an Information Retrieval (IR) system. The person typing it has
  \textbf{NOT} seen this document and is searching to find it. Never refer to
  the document or its contents. This includes not only overt phrases such as
  ``described in the invention'', ``according to the text'', or ``in the
  passage'', but also any wording that presupposes the reader is looking at the
  document, such as ``the example given'', ``the approach
  mentioned/described/disclosed here'', or ``which method is provided''. Write
  self-contained questions. Bad: ``Which agent example is given that increases
  ROS?'' Good: ``Which agent increases ROS?''

  \item \textbf{Single Intent:} Each question has one clear informational goal.
  It is fine if the goal is broader than a single fact, for example ``how does
  X address Y'', but it should not bundle multiple unrelated asks.

  \item \textbf{Diversity:} The three questions must target three
  distinguishable aspects of the passage. They must span at least two of the
  following framings: problem, solution, and application.

\end{itemize}

\medskip
\textbf{FRAMINGS}

\begin{enumerate}[leftmargin=1.5em, itemsep=0pt, topsep=1pt]
  \item \textbf{Problem-framed:} what challenge or limitation does the passage
  address?
  \item \textbf{Solution-framed:} what approach, mechanism, or method is
  proposed?
  \item \textbf{Application-framed:} what use case, scenario, or capability
  does the passage enable?
\end{enumerate}

Broader conceptual questions about the overall approach are acceptable as long
as the passage clearly and directly answers them in a contiguous span.

\medskip
\textbf{EXAMPLES}

\textbf{Example 1}

\textbf{Passage (Turkish, ASCII-normalized):}

Enerji tuketimi dusurulmus bir tahrik tertibati. Bulus hareket enerjisi
uretebilen bir motorun (10) en az bir yanma odasinda (11) ihtiyac duydugu
kimyasal unsurlardan en azindan bir kisminin temin edilerek enerji sarfiyatinin
azaltilabildigi bir tahrik tertibati (1) ile ilgilidir. Bulusun yeniligi;
bahsedilen tahrik tertibatinin (1) en az bir elektroliz birimi (20) ve en az
bir kontrol birimi (40) ile iliskilendirilmis olmasi, bahsedilen elektroliz
biriminin (20) elektrolitin elektroliz edilmesini saglayacak sekilde konfigure
edilmis olmasi, bahsedilen kontrol biriminin (40) ise elektroliz biriminde
bulunan elektroliti elektroliz edilirken yaklasik 85--87$^\circ$C araligina
kadar isitacak sekilde gerilim ve voltaji ayarlayabilir olmasi, elektroliz
biriminde (20) elektroliz sonrasinda elde edilen oksijen, hidrojen ve su
buharinin bahsedilen yanma odasina (11) karistirilarak verilebilir olmasidir.

\textbf{Good, application-framed, paraphrased:} ``How can an engine reduce fuel
consumption by generating part of its own combustion inputs?''

This captures the core concept of the passage. It uses ``generating part of its
own combustion inputs'' instead of lifting phrases such as ``electrolysis'' and
``combustion chamber''. The answer points to the drive arrangement producing
oxygen, hydrogen, and water vapor and feeding them into the combustion chamber.

\textbf{Good, solution-framed, paraphrased:} ``What role does on-board
electrolysis play in supplying reactants to an engine?''

``Electrolysis'' is a technical term that stays. ``Supplying reactants to an
engine'' paraphrases the passage content about chemical components needed by
the engine and delivery to the combustion chamber.

\textbf{Bad, too extractive and keyword-heavy:} ``What temperature range is
maintained for the electrolyte during electrolysis?''

This is a fact-extraction question, not a semantic retrieval question. It
belongs to the technical prompt, not this one.

\textbf{Bad, ungrounded conceptual:} ``What are the environmental benefits of
this fuel-saving approach?''

This sounds like a retrieval question, but the passage does not discuss
environmental benefits. Discard.

\medskip
\textbf{Example 2}

\textbf{Passage (French, ASCII-normalized):}

Systeme de surveillance de traitement d'eau decentralise connecte a un reseau
et dispositif de suivi de fonctionnement et de maintenance comprenant au moins
un capteur connecte a Internet et une plateforme logicielle integree qui peut
etre utilisee par des installations de grande taille et/ou des organisations
comprenant de multiples installations qui gerent et maintiennent des systemes
de traitement d'eau potable decentralises, par exemple des filtres de point
d'utilisation, ou des processus de traitement d'eau potable decentralises, par
exemple un rincage de robinet, a un ou a plusieurs emplacements. Au moins un
capteur est installe en amont ou en aval d'un consommable ou d'un processus de
traitement d'eau decentralise et mesure des donnees d'eau, comprenant non
exclusivement le volume d'eau traite dans le temps, et alerte l'utilisateur
lorsque des actions de maintenance sont requises pour maintenir la qualite de
l'eau.

\textbf{Good, problem-framed, paraphrased:} ``How can organizations with
multiple drinking water sites know when filters need servicing?''

This captures the problem the passage addresses using natural search wording.
It avoids lifting terms such as ``consommable'' or ``actions de maintenance''.
The answer points to the sensor alerting the user when maintenance is required.

\textbf{Good, application-framed:} ``Can IoT sensors monitor point-of-use water
filters across many locations?''

``IoT sensors'' paraphrases ``capteur connecte a Internet''. ``Point-of-use
water filters'' is a reused standard term and is acceptable. The answer is
grounded in the passage description of the monitoring system.

\textbf{Bad, too extractive:} ``What does the sensor measure in the
decentralized water treatment system?''

This is a fact-extraction question. It belongs to the technical prompt.

\textbf{Bad, keyword-lifted:} ``What is a decentralized water treatment
monitoring system connected to a network?''

This is near-verbatim reuse of the passage opening phrase. It would match via
keyword overlap, defeating the purpose of semantic retrieval.

\medskip
\textbf{SELF-CHECK}

For each question before finalizing, check:

\begin{itemize}[leftmargin=1.2em, itemsep=0pt, topsep=1pt]

  \item Does the question ask about a concept, problem, approach, or
  application, rather than a single extracted fact?

  \item Is the question vocabulary meaningfully different from the passage
  vocabulary while preserving meaning? Standard technical terms and proper
  nouns are exempt.

  \item Can you point to a contiguous span in the passage that directly answers
  it? If not, discard.

  \item Would someone searching for a document like this one plausibly type
  this query without already having read it?

  \item Could someone who has never seen this document type this question into
  a search engine? It must not refer to the document, an ``example given'', or
  anything ``mentioned'' or ``described'' in it.

  \item Do the three questions cover at least two framings: problem, solution,
  and application?

\end{itemize}

\medskip
\textbf{OUTPUT:} valid JSON only, no markdown.

\begin{lstlisting}[style=jsonstyle]
[
  {"question": "...", "answer": "...", "framing": "..."},
  {"question": "...", "answer": "...", "framing": "..."},
  {"question": "...", "answer": "...", "framing": "..."}
]
\end{lstlisting}

\texttt{answer}: a short contiguous span taken from the passage that directly
answers the question. Keep it as short as possible while fully addressing the
question. The answer must be in English.

\texttt{framing}: one of \texttt{problem}, \texttt{solution}, or
\texttt{application}.

The same passage exists in different languages. This should not affect your
target language for questions and answers. They are declared above. Here are
the passages:

\end{tcolorbox}

\vspace{0.5em}

\captionof{figure}{General-question generation prompt, English version.
Language-specific variants differ only in the declared output language.}
\label{fig:prompt-general}

\vspace{1em}

\begin{center}
  \includegraphics[width=\textwidth]{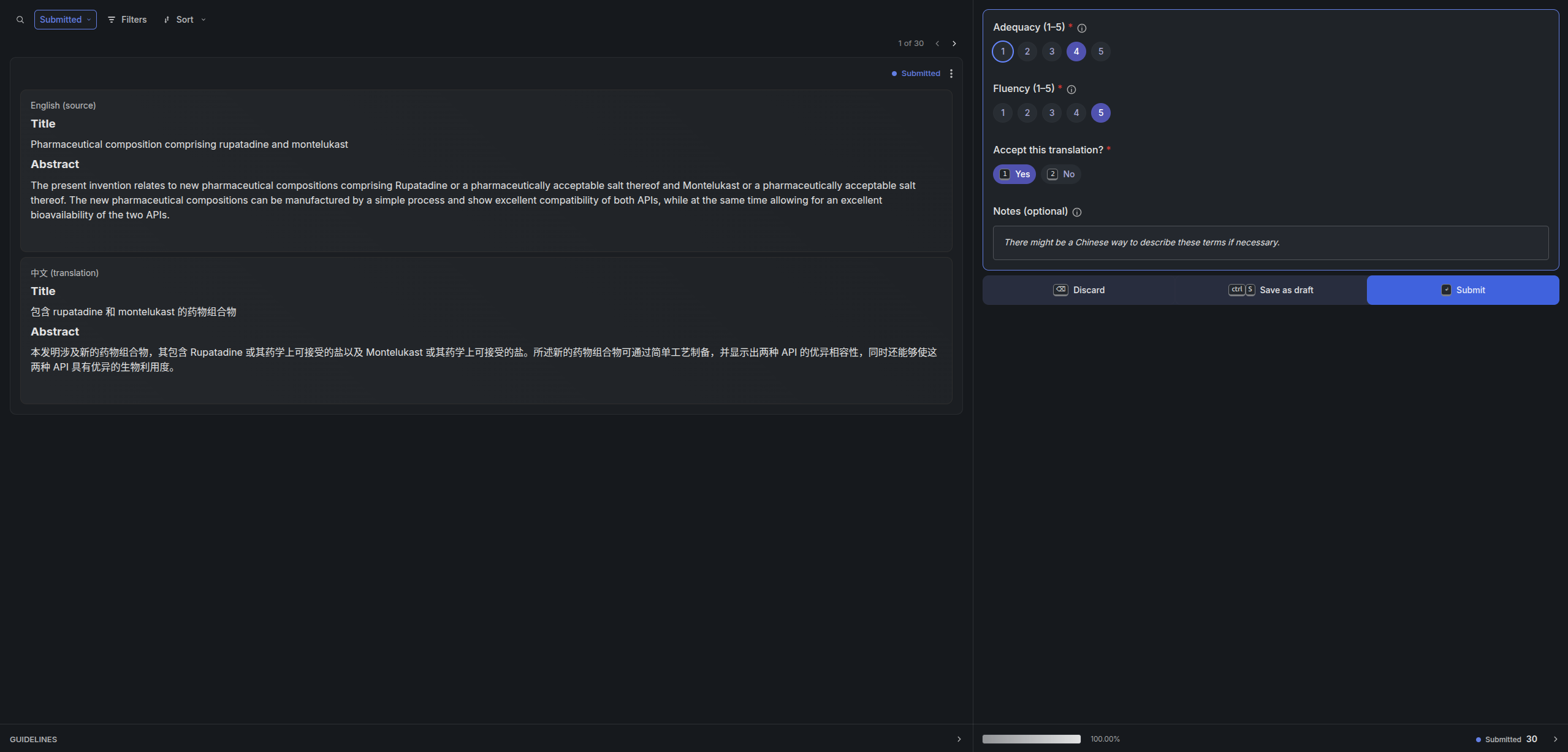}

  \vspace{0.5em}

  \captionof{figure}{Human annotators use the Argilla open-source tool to perform annotation tasks.}
  \label{fig:argilla-annotation}
\end{center}